\documentclass[times,review,10pt]{elsarticle}

\usepackage{lineno,hyperref}
\modulolinenumbers[5]
\usepackage{amsmath}
\usepackage{colortbl}
\usepackage{listings}
\usepackage{multirow}
\usepackage{makecell}
\usepackage{enumitem}

\usepackage{xcolor}
\usepackage{soul}
\usepackage[utf8]{inputenc}
\usepackage[small]{caption}
\usepackage{multirow}
\usepackage{comment}
\usepackage{booktabs}
\usepackage{graphicx}
\usepackage{graphics}
\usepackage[export]{adjustbox}
\usepackage{bm}
\usepackage{amsmath,amsfonts}
\journal{Journal of \LaTeX\ Templates}

\usepackage{colortbl}
\definecolor{citecolor}{RGB}{119,185,0} 
\usepackage[dvipsnames,svgnames,x11names]{xcolor}
\usepackage[ruled,linesnumbered]{algorithm2e}
\usepackage{amssymb}
\usepackage{subfig}
\usepackage{tabularx}
\usepackage{array}

\newlength\savewidth
  
\def\eg{\emph{e.g.}} 
\def\ie{\emph{i.e.}} 
\def\etal{\emph{et~al.}} 

\begin{document}

\begin{frontmatter}



\title{Harnessing Weak Pair Uncertainty \\ for Text-based Person Search}




\author[mymainaddress]{Jintao Sun}
\ead{3120215524@bit.edu.cn}

\author[mysecondaryaddress]{Zhedong Zheng}
\ead{zhedongzheng@um.edu.mo}

\author[thirdaddress]{Gangyi Ding}
\ead{dgy@bit.edu.cn}

\address[mymainaddress]{School of Computer Science and Technology, Beijing Institute of Technology, China}
\address[mysecondaryaddress]{Faculty of Science and Technology, University of Macau, China}
\address[thirdaddress]{School of Computer Science and Technology, Beijing Institute of Technology, China}

\begin{abstract}
In this paper, we study the text-based person search, which is to retrieve the person of interest via natural language description. Prevailing methods usually focus on the strict one-to-one correspondence pair matching between the visual and textual modality, such as contrastive learning. However, such a paradigm unintentionally disregards the weak positive image-text pairs, which are of the same person but the text descriptions are annotated from different views (cameras). To take full use of weak positives, we introduce an uncertainty-aware method to explicitly estimate image-text pair uncertainty, and incorporate the uncertainty into the optimization procedure in a smooth manner. Specifically, our method contains two modules: uncertainty estimation and uncertainty regularization. (1) Uncertainty estimation is to obtain the relative confidence on the given positive pairs; (2) Based on the predicted uncertainty, we propose the uncertainty regularization to adaptively adjust loss weight. Additionally, we introduce a group-wise image-text matching loss to further facilitate the representation space among the weak pairs. Compared with existing methods, the proposed method explicitly prevents the model from pushing away potentially weak positive candidates. Extensive experiments on three widely-used datasets, \ie, CUHK-PEDES, RSTPReid and ICFG-PEDES, verify the mAP improvement of our method against existing competitive methods +3.06\%, +3.55\% and +6.94\%, respectively.
\end{abstract}




\end{frontmatter}




\section{Introduction}
\label{sec:intro}
Text-based person search is an extension of conventional image-based person re-identification (re-ID)~\citep{LIU2022108654,Shu_Wen_Wu_Chen_Song_Qiao_Ren_Wang_2022}, which is to retrieve the person of interest from a large pool of candidate images given text descriptions. In real-world scenarios, the image query of the target person usually is not accessible, while the test description is easy to obtain~\cite{APTM,oursMM,KE2024110481}. 
Therefore, more researchers resort to the text-based person search. The key underpinning this task is to mine the fine-grained information of images and texts, and establish the cross-modality alignment. 

With the advancement of cross-modality learning, numerous deep learning approaches have been proposed, which can be broadly categorized into two directions. The first direction focuses on data augmentation, primarily by generating additional data and employing the pretrain-finetune paradigm. 
For example, based on a pre-trained model, Jiang and Ye~\cite{jiang2023crossmodal} propose a momentum distillation cross-modal method using four datasets for pre-training, which enables the model to utilize larger noisy datasets, thereby improving learning under noisy supervision. 
Differently, Yang~\etal~\cite{APTM} propose a large-scale image-text dataset with high similarity to the target dataset, constructed using a diffusion model, which addresses the challenges of image collection and time-consuming text annotation.
The second direction emphasizes metric learning, aiming to design more effective loss functions to better exploit multimodal information within the data, such as contrastive loss and cross-modal matching loss~\cite{ALBEF}. 
For instance, Shao~\etal~\cite{Shao_Zhang_Fang_Lin_Wang_Ding_2022} propose to use cross entropy loss and ranking loss to get the total loss of multimodal shared storage dictionaries. Bai~\etal~\cite{Bai_2023} incorporate relation-aware loss and sensation-aware loss, enabling the model to focus more on the details of image-text pairs and learn cross-modal features with finer granularity. 

\begin{figure}[t!]
  \centering
  \includegraphics[width=0.98\linewidth]{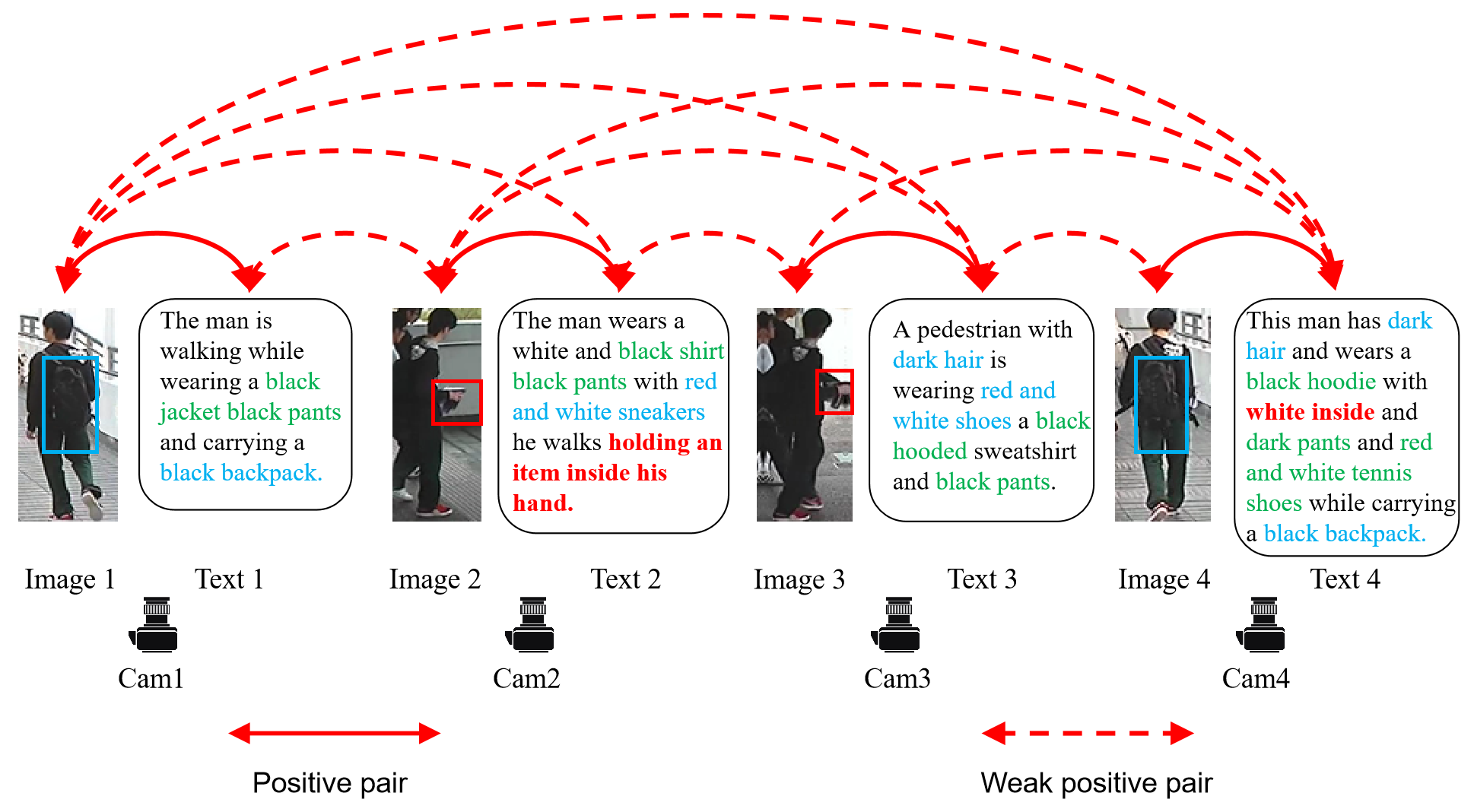}
\vspace{-.1in}
  \caption{Here we show the discrepancy between weak positive pairs (dotted arrows) and positive pairs (solid arrows) of the same identity. For instance, there are four image-text pairs of one person. {\color{ForestGreen}{\textbf{Green}}} shows a shared description among four texts, {\color{cyan}{\textbf{blue}}} denotes the description discrepancy, and {\color{red}{\textbf{red}}} indicates the unique description. 
  We can observe that the text description is strongly related to camera views instead of only depending on the identity. Such difference is mainly due to the text annotation process, where annotators can only see a single view of the person. 
  Considering such matching difficulty, most previous methods, thus, take advantage of the positive pairs for strict one-to-one matching (solid arrows), while disregarding the weak positive pairs (dashed arrows). In contrast, we mine the relation among weak positive image-text pairs and explicitly harness such weak positive pairs to learn discriminative cross-modality matching. 
  }
  \label{fig:1}
\end{figure}

We observe that there is an inherent limitation in text-based person retrieval metric learning, which primarily relies on strict one-to-one contrastive learning on positive pairs. 
The weak positive pairs are usually disregarded due to the annotation discrepancy. The annotator only observes one view (camera angle) of the target person, and can not provide a comprehensive description. 
As illustrated in Fig.~\ref{fig:1}, images of the same person usually be annotated with variations in local details, resulting in a mismatch between the image and text descriptions across different perspectives. Consequently, strict one-to-one matching discards these weak positive pairs during training, overlooking potentially valuable information.
However, we argue that weak positive pairs play a crucial role in enhancing the model ability to capture shared patterns across different views of the same individual. These shared patterns should be positioned closer to the anchor points in the feature space than negative pairs, enabling the model to better generalize across varying perspectives. Despite this, the potential of weak positive relationships remains under-explored in current methodologies.

To take full usage of weak positive pairs from the dataset, we propose a new uncertainty-aware learning method to solve some unutilized data problems. There are two steps in total. First, we adjust uncertainty for the comparison learning of data features that can be most affected by data underutilized. Under the condition of keeping a one-to-one correspondence comparison of positive data features, we leverage the auxiliary information of weak positive pairs. The feature similarity of images and texts of weakly positive pairs are calculated respectively, and parameters are set to harness and adjust the feature contrast of weakly positive pairs, which is added to the total loss calculation, thus improving the model ability to learn from data. In the second step, we exploit the impact of data representation space in the construction of negative pairs during metric learning to both increase the quantity and difficulty of negative pairs. Additionally, our group-wise matching method makes full use of the information of weak positive pairs to make the representation space distribution of the model more reasonable. This approach enables the calculation of the matching loss for 1 pair of positive samples, 2 pairs of weak positive samples, and 6 pairs of negative samples. Finally, our uncertainty-aware approach greatly improves the learning accuracy of the model without adding additional modules.
In summary, our contributions are: 

\begin{itemize}[leftmargin=*,labelsep=0.8em]
\item  We observe a limitation in the existing text-based person search training, which stems from the strict one-to-one correspondence contrastive learning approach. To address this issue, we propose an uncertainty learning-based method that effectively leverages underutilized weak positive pairs. Specifically, our method incorporates uncertainty into the feature comparison process, enabling the model to leverage weak positive pairs rather than discarding them. As a minor contribution, we also propose a Group-wise Image-Text Matching (GITM) loss, which facilitates the matching of weak positive pairs in a group-wise manner.
\item 
Extensive experiments verify that our uncertainty-aware method, considering weak positive pairs during training, recalls more positive candidates to the top ranking. In particular, our approach outperforms competitive methods, \eg, RaSa and APTM, by 3.06\%, 3.80\%, 6.94\% mAP and 5.53\%, 3.55\%, 7.01\% mAP on CUHK-PEDES, RSTPReid, and ICFG-PEDES, respectively.
\end{itemize}


\section{Related Work}\label{related_work}
\subsection{Text-Image Person Search}
Text-based person search constitutes a challenging fine-grained cross-modal retrieval task, prompting the emergence of diverse methodologies in recent years. Existing approaches are primarily categorized into two paradigms: those leveraging explicit cross-modal attention interactions to enhance regional-word/phrase correspondence and predict image-text matching scores through sophisticated attention mechanisms~\cite{Wang_Zhu_Xue_Wan_Liu_Wang_Li_2022, ZHANG2025111247, LIU2023109636, Shu_Wen_Wu_Chen_Song_Qiao_Ren_Wang_2022}, which improve inter-modal fusion at the cost of elevated computational complexity, and lightweight alternatives that forgo such interactions by learning aligned representations within a shared feature space via carefully designed architectures and objectives~\cite{Chen_Zhang_Lu_Wang_Zheng_2022, Zheng_Zheng_Garrett_Yang_Xu_Shen_2020}. Early efforts, such as dual-path convolutional frameworks~\cite{Zheng_Zheng_Garrett_Yang_Xu_Shen_2020}, exploit end-to-end supervision to derive modality-specific features, while more recent advances incorporate vision-language pre-training to yield robust representations~\cite{Shu_Wen_Wu_Chen_Song_Qiao_Ren_Wang_2022, APTM, Bai_2023}, often augmented by attribute prompt learning, relation-aware modeling, or multi-attribute datasets like MALS to facilitate fine-grained alignment. Nevertheless, prevailing methods largely overlook the rich auxiliary supervisory signals embedded in weak positive pairs within the datasets. In contrast, the proposed approach introduces uncertainty learning and regularization to refine both contrastive learning and image-text matching objectives, thereby fully harnessing weak positive information to enhance the discriminative capacity of positive pair representations.
    
\subsection{Uncertainty Learning}
Uncertainty quantification has become increasingly important in data-driven methods as datasets grow larger and demands for model reliability intensify. Kendall and Gal~\cite{Kendall_Gal_2017} provide a foundational taxonomy that decomposes predictive uncertainty into aleatoric uncertainty, which captures irreducible data-inherent noise, and epistemic uncertainty, which reflects model parameter ambiguity arising from limited or insufficient training data and can be reduced through additional observations or targeted refinements. Aleatoric uncertainty has been extensively explored in computer vision tasks, including image retrieval~\cite{Warburg_Jorgensen_Civera_Hauberg_2021,chen2024composed}, classification~\cite{Postels_Segu_Sun_Gool_Yu_Tombari_2021}, and segmentation~\cite{10.1007/s11263-020-01395-y}, with approaches such as explicit noise injection into features~\cite{Chang_Lan_Cheng_Wei_2020,dou2022reliabilityaware}, Monte Carlo estimation of distributional similarity~\cite{oh2019modeling}, and loss-variance-aware reweighting~\cite{9710535}. In contrast, epistemic uncertainty modeling commonly leverages Bayesian frameworks~\cite{Marchette_2003,Gal_Ghahramani_2015}, with Monte Carlo Dropout~\cite{Gal_Ghahramani_2015} serving as a practical approximation by introducing stochasticity during inference; recent works further integrate cross-modal biases~\cite{jiang2023crossmodal} and dynamic uncertainty-guided learning. Multi-granularity annotation strategies combined with Gaussian noise simulation have also been employed to explicitly address aleatoric effects in composed image retrieval~\cite{chen2024composed}. Building on the insights from previous work, this study fully leverages the information contained in weak positive pairs within the dataset. 
There are two fundamental differences between previous work and our approach: (1) We do not introduce additional modules or parameters to simulate uncertainty using noise. Instead, we harness the weak positive pairs, which describe different image-text pairs for the same ID, as a source of uncertainty to support the model contrastive learning of image-text features. (2) We employ uncertainty regularization, and apply a group-wise strategy to incorporate the semantic information from weak positive pairs, thereby enhancing the image-text matching process.

\begin{figure*}[ht!]
  \centering
  \includegraphics[width=\linewidth]{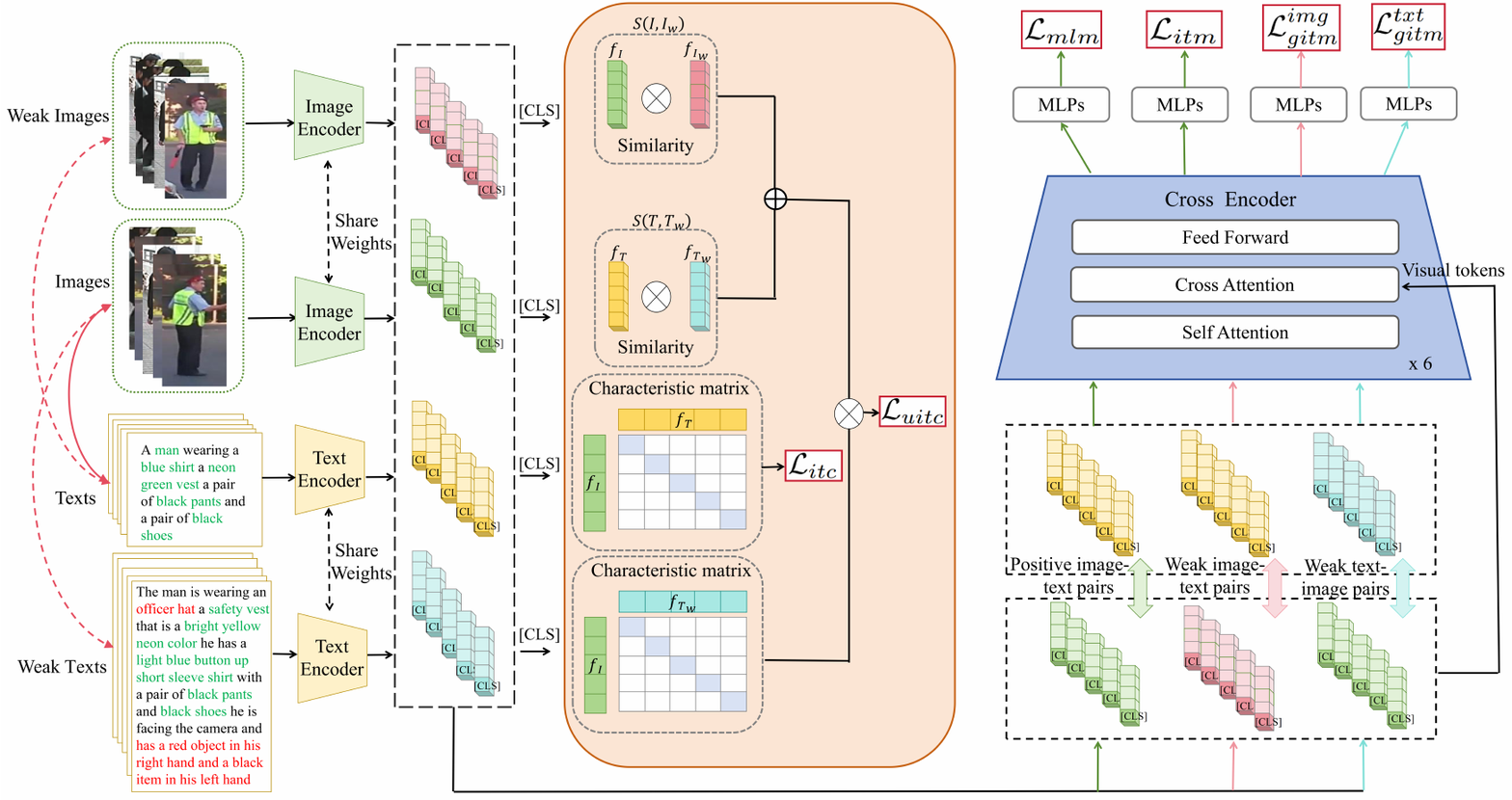}
  \vspace{-1.2in}
  \caption{ An architecture overview of our approach. Firstly, weakly positive text and weakly positive image corresponding to the anchor image and text are randomly selected from the dataset according to the same ID. Then we send all sampled data to the image / text encoder to obtain the corresponding features. 
  Secondly, we compute the ITC loss between images and texts. 
  For the anchor image-text pairs, we calculate the original contrastive learning loss, \ie, $\mathcal{L}_{itc} (I,T)$ via the aggregated embeddings of [CLS]. For the weak image-text pairs, we derive the contrastive objective, \ie, $\mathcal{L}_{itc} (I,T_w)$, and further regularize this term with uncertainty estimation as $\mathcal{L}_{uitc}$.
  The proposed uncertainty-aware contrastive loss $\mathcal{L}_{uitc}$ adaptively adjust the metric learning between weakly positive texts and images.
  Thirdly, ITM loss and MLM loss are calculated based on the features of image-text pairs. For weak positive counterparts, 
  we introduce Group-wise Image-Text Matching (GITM) loss to facilitate the representation learning. 
  }
  \label{fig:2}
\end{figure*}

\section{Methodology} \label{proposed method}

\subsection{Preliminaries}~\label{sec:3.1}
We employ 
the prevailing APTM framework~\cite{APTM} as our baseline model. This framework comprises two primary phases: pre-training on a synthesized dataset and fine-tuning on downstream datasets. During the pre-training phase, Attribute Prompt Learning (APL) and Text Matching Learning (TML) are employed to capture shared knowledge relevant to text-based person search and pedestrian attribute recognition. In the fine-tuning phase, downstream datasets are utilized to further optimize the model parameters. 
The baseline comprises three encoders: image encoder, text encoder, and cross encoder, along with two MLPs-based headers. In this work, we do not change the pre-training phase and only apply our uncertainty estimation and uncertainty regularization methods in the finetune phase.
We do not pursue the network contribution in this work.
Therefore, we adopt the common backbone for a fair comparison. The influence of the original image encoder and the replaced image encoder on the results are compared in detail in the ablation studies.
The $\left [ \mathrm{CLS}  \right ] $ embedding represents the entirety of the image / text.
The cross-encoder integrates image and text representations for prediction tasks, thereby discerning their semantic relationship.

\subsection{Network Structure}\label{sec:3.2}

In this paper, we do not pursue the deployment of complex network structures but instead offer a new learning strategy. We mainly follow the existing work~\cite{APTM,oursMM} to construct the network for a fair comparison. Given that datasets contain images and text descriptions of the same person from different perspectives, previous work often overlooks the weak correlation between images and text under the same ID. However, the auxiliary information of weak positive pairs can significantly enhance the model ability to learn a more comprehensive representation of the description of person.
Therefore, we mainly consider the uncertainty of image-text pairs describing the same person in the dataset. As shown in Fig.~\ref{fig:1}, the dataset usually contains multiple image-text pairs describing the same ID. The previous method relies solely on one-to-one image-text pairs for training, overlooking the relationship between the current image and the texts captured from another perspective, as well as the connection between the current text and images taken from different angles.
Consider that these are not exactly the same texts and images. If the text is biased relative to the source image, or if the target image contains more perspective information than the source text, the system will supplement this uncertain information. In the training stage, in order to reduce the visual information bias and text description bias of the same person from different perspectives, in this work, we mainly study the uncertainty in cross-modal data matching.
We show a brief pipeline in Fig.~\ref{fig:2}. A one-to-one correspondence positively correlated image-text pair $\left( I, T \right)$ and weakly correlated image $ I_{w} $ and text $ T_{w} $ randomly selected for this image-text pair $\left( I, T \right)$ are extracted from the dataset. Our model extracts the features of image $I$ to obtain $ f_{I} $ and $ f_{I_{w}} $. Text encoder extracts text features $T$ to get $f_{T}$ and $f_{T_{w}}$, and carries out contrastive learning for image-text features. Meanwhile, weak positive pairs uncertainty are added to the contrastive learning. The obtained image embeds and text embeds, as well as hard negative pairs obtained by contrast learning based on image text features and hard negatives, increased based on our group-wise method, are sent to cross encoder together. The specific cross-modal data uncertainty approach is detailed in the following two sections.
 
\subsection{Uncertainty Estimation} \label{sec:3.3}

\textbf{Motivations.} Existing text-based person search datasets provide image and text descriptions from different perspectives for the same individual. However, current methods often focus solely on one-to-one image-text pairs, which frequently fail to fully capture the characteristics of a person due to feature deviations across different perspectives. Learning features from strictly one-to-one image-text pairs can be limiting, as the method only leverage one-to-one image-text pairs usually overlook certain characteristics, leading the model to perform retrieval based on incomplete information.

First, we define the one-to-one corresponding image-text pairs $\left( I, T \right)$ of the input model in the dataset as positive pairs, the unmatched image-text pairs as negative pairs, and the weakly positive correlation image-text pairs $\left( I_{w}, T_{w} \right)$. $I_{w}$ is the weakly positive image relative to $T$ obtained by random extraction according to the current one-to-one corresponding image-text pair $\left( I, T \right)$ in the case of describing the same person (same ID), and $T_{w}$ is the weakly positive text description relative to $I$ obtained by random extraction under the case of the same ID. The existing methods usually adopt Image-Text Contrastive Learning (ITC) to distinguish positive and negative pairs. Given a matched pair $\left( I,T \right)$ we initially extract their respective representations $f_{I}$ and $f_{T}$. We denote the set of all matched image-text pairs in a mini-batch as $B$. The matching score can be simply formulated as:
\begin{equation}
  S\left ( I, T\right ) =\frac{\mathrm{exp} \left ( \cos\left ( f_{I},f_{T} \right ) /\tau  \right ) }{ { {\textstyle \sum_{i=1}^{B}}  \left ( \mathrm{exp} \left ( \cos\left (  f_{I},f_{T^{i} } \right ) /\tau  \right ) \right )  } } ,
 \label{eq:1}
\end{equation}
where $\tau $ is a learnable temperature parameter, $\cos(\cdot,\cdot)$ means the cosine similarity, and $\mathrm{exp}$ denotes the exponential function. Similarly, given the text and a batch of images, we calculate the matching score of the paired image $S\left ( T, I\right )$. The ability to differentiate learning is added to the final loss calculation. The ITC loss is defined as follows:
\begin{equation}
\mathcal{L} _{itc}\left ( I, T\right )=-\mathbb{E}  \left ( \mathrm{log}S\left ( I, T\right ) + \mathrm{log}S\left ( T, I\right ) \right ) .
 \label{eq:2}
\end{equation}

We compute cosine similarity on L2-normalized embeddings (as in standard retrieval baselines), hence $\cos(\cdot,\cdot)\in[-1,1]$.
Moreover, the ITC objective in Eq.~(\ref{eq:1}) already includes a temperature parameter $\tau$ in the logits ($\cos(\cdot,\cdot)/\tau$), which controls the sharpness of the softmax distribution and stabilizes gradients; $\tau$ is learnable in our implementation, consistent with the baseline. 

In order to take full advantage of the annotation, we propose an uncertainty estimation method. We integrate the weak positive pair into the text image alignment. Since our task is text-based image retrieval, the first step of our uncertainty estimation method is to extract the features $f_{I}$ of the positive image $I$ and the text feature $f_{T_{w}}$ of the weak text $T_{w}$ is with the same ID but annotated based on a different viewpoint. 
Then we calculate the obtained image feature $f_{I}$ and weak text feature $f_{T_{w}}$ matching score, refer to Eq.~\ref{eq:1}. Therefore, we can obtain the matching score $S\left ( I, T_{w}\right )$ of positive image features and weak positive text features with the same ID. Similarly, we can get the matching score $S\left ( T, I_{w}\right )$ of positive text features and weak positive image features with the same ID. According to the conventional ITC method Eq.~\ref{eq:2}, $\mathcal{L} _{itc} \left ( I,T_{w}  \right ) $ is calculated for subsequent uncertainty regularization. Next, we extract the corresponding image and text features $f_{I_{w}}$ and $f_{T_{w}}$ according to the input weak positive pair $\left( I_{w}, T_{w} \right)$. The image feature $f_{I}$ and the text feature $f_{T}$ of the positive image-text pair $\left(I, T \right)$ with the same ID. Then, we calculate the similarity between $f_{I}$ and $f_{I_{w}}$ , $f_{T}$ and $f_{T_{w}}$ respectively. We define $\cos\left (  f_{I},f_{I_{w} } \right )$ as the similarity calculated between $f_{I}$ and $f_{I_{w}}$. Similarly, $\cos\left (  f_{T},f_{T_{w} } \right )$ is derived in the same manner. We define the consistency score $s_{w}$ as the sum of intra-modality similarities between the anchor and weak positive samples. To measure the semantic discrepancy across views, we compute the uncertainty $u_{w}$ via an exponential transform:
\begin{equation}
s_w=\frac{1}{2}\Big(\cos(f_I,f_{I_w})+\cos(f_T,f_{T_w})\Big),
\qquad
u_w=exp\left(-s_w\right).
\label{eq:sim}
\end{equation}
Here, $f_I$ and $f_T$ denote the image/text features of an anchor sample, while $f_{I_w}$ and $f_{T_w}$ are the corresponding features of its paired sample $w$ (e.g., another view or a weakly matched counterpart). $\cos(\cdot,\cdot)$ is the cosine similarity. Since cosine similarity is bounded in $[-1,1]$, we have $s_w\in[-1,1]$ and thus $u_w=\exp(-s_w)\in[e^{-1},e^{1}]$.
Therefore, $u_w$ is strictly positive and bounded, and larger $u_w$ indicates lower cross-view consistency (higher ambiguity) in weak-pair matching. Our uncertainty estimation method fully leverages auxiliary information of images and texts and utilizes annotation information in data sets to supplement the information between images and texts, so as to enhance the model ability to accurately capture and interpret the nuanced relationships between visual and textual data, leading to more precise and reliable feature extraction and matching.


\textbf{Discussion. What is the advantage of uncertainty estimation in feature learning?}  
Uncertainty estimation is pivotal in enhancing the robustness of feature learning by simulating scenarios in which different views or descriptions of the same person are matched with alternative views or textual descriptions under the same ID. This approach effectively mitigates discrepancies that commonly arise between varying descriptions of the same individual, thereby reducing deviations that can negatively impact model performance. Moreover, uncertainty estimation addresses the challenge of overfitting associated with strict one-to-one matching, enabling the model to generalize more effectively and to reason in a manner more aligned with human cognition.
By diminishing the homogeneity and reducing the deviations between image and text descriptions, uncertainty estimation enhances the model ability to learn more comprehensive and accurate representations. This improvement not only bolsters the model robustness against variations in descriptions but also leads to superior performance in real-world scenarios, where perfect alignment between descriptions and images is often lacking. Our experimental results verify that the proposed method significantly outperforms existing approaches, particularly in terms of mean Average Precision (mAP), highlighting its effectiveness in addressing the complexities inherent in text-based person search tasks.

\textbf{Applicability and limitations.}
The proposed uncertainty learning is most effective when weak positives are informative but imperfectly aligned, as in TBPS datasets with multi-view annotations.
It may yield limited gains when weak positives become noisy supervision (e.g., severe occlusion/viewpoint changes, non-overlapping attribute descriptions, or annotation mismatch), where a weak positive can behave close to a pseudo-negative in the contrastive space.
In practice, Fig.~\ref{fig:uncertainty_reliability} provides a diagnostic: high-$u$ queries/pairs exhibit higher retrieval risk, and a large mass of high-$u$ weak pairs indicates that weak-positive supervision is less reliable.

\subsection{Uncertainty Regularization} \label{sec:3.4}

\textbf{Motivations.} Since conventional ITC and ITM loss only rely on one-to-one correspondence positive pairs in the data set, and do not maximize the utilization of the weak positive information in the labeling, in order to make full utilize of this information to assist us in training the model, we propose uncertainty regularization to adjust the model learning of ITC loss, and group-wise metric learning for ITM loss so that the model can better grasp the information in the data. The existing work aims at negative pair mining in ITM loss. The conventional ITM method only compares and learns $\left( I, T \right)$ based on one-to-one correspondence mapping of images and texts in the input dataset to obtain negative pair construction. Such construction will ignore some useful auxiliary weak positive instances and have less structural information. The difficulty of the negative pair is only determined by the one-to-one correspondence image-text pair and the batch size, and much auxiliary information is not considered in the weak positive pair, so the difficulty of the negative pair has certain limitations. Moreover, in existing works, the ratio of positive pair to negative pairs is fixed at 1:2, where for each input image-text pair $\left( I, T \right)$, the image is used to find a negative text, and the text is used to find a negative image. Consequently, the number of negative pairs is fixed, which limits the quantity and diversity of negative pairs available.

To solve the above problem, in particular, we exploit to the fullest extent the image features and text features of weak positive pairs in the task and obtain the loss of images and weak positive texts through uncertainty estimation in Section~\ref{sec:3.3}.
The similarity between the extracted image and text features of positive pairs and those of weak positive pairs with the same ID, denoted as $\cos\left (  f_{I},f_{I_{w} } \right )$ and $\cos\left (  f_{T},f_{T_{w} } \right )$, respectively, we can obtain an uncertainty-aware ITC loss: 
\begin{align}
\mathcal{L} _{uitc}   = \frac{ \mathcal{L} _{itc} \left ( I,T_{w}  \right )  }{\gamma \times u_{w} }  + \gamma\times u_{w} ,
\end{align}
where $\gamma$ is a learnable parameter used to dynamically adjust the uncertainty adjustment, and $\mathcal{L} _{itc} \left ( I,T_{w}  \right ) $ is calculated by inputting image feature $f_{I}$ and with the text feature $f_{T_{w}}$ of the weak positive of this image into the original ITC loss calculation and applying Eq.~\ref{eq:2} to calculate. 
The uncertainty $u_{w}$ is derived via an exponential mapping of the average intra-modality similarities between the weak positive samples and the anchor pairs, as formulated in Eq.~\ref{eq:sim}.
In implementation, we stop the gradient through $u_w$ when applying uncertainty regularization, so that $u_w$ serves only as a detached reliability weight and the network cannot reduce Eq.~(4) by directly manipulating the weighting path.

We note that ITC loss only focuses on the one-to-one matching of image text pairs, and as we mentioned earlier, these one-to-one correspondence text pairs are not comprehensive when describing a person. Therefore, in this work, inspired by uncertainty, we propose an uncertain regularization to optimize the existing ITC loss. In fact, the existing ITC loss is a special case of our uncertainty adjustment, when the input only has a one-to-one correspondence text pair corresponding to one person and no other multiple descriptions or perspectives.

 Image-text Matching (ITM) Learning is a binary classification method for predicting whether an input image and text match. Eq.~\ref{eq:1} in ITC is used to calculate the similarity of input images and text features and select the unpaired image with the highest similarity to each text as the hard negative. Similarly, we could select the unpaired text with the highest similarity as the hard negative for images. Such a pair of positive samples and two pairs of negative samples go through the cross-encoder to obtain ITM loss as follows:
 \begin{align}
\mathcal{L}_{itm} = \mathbb{E} \left [ p \log\hat{p} (I,T)  +\left (  1-p \right ) \log\left (1- \hat{p} (I,T)\right ) \right ],
\label{eq:4}
\end{align}
where $p$ is 1 if $(I, T) $ is matched, 0 otherwise, and $\hat{p}$ is the estimated match score of image-text pairs calculated by an MLPs with Sigmoid activation.

\begin{figure}[t!]
  \centering
  \includegraphics[width=0.95\linewidth]{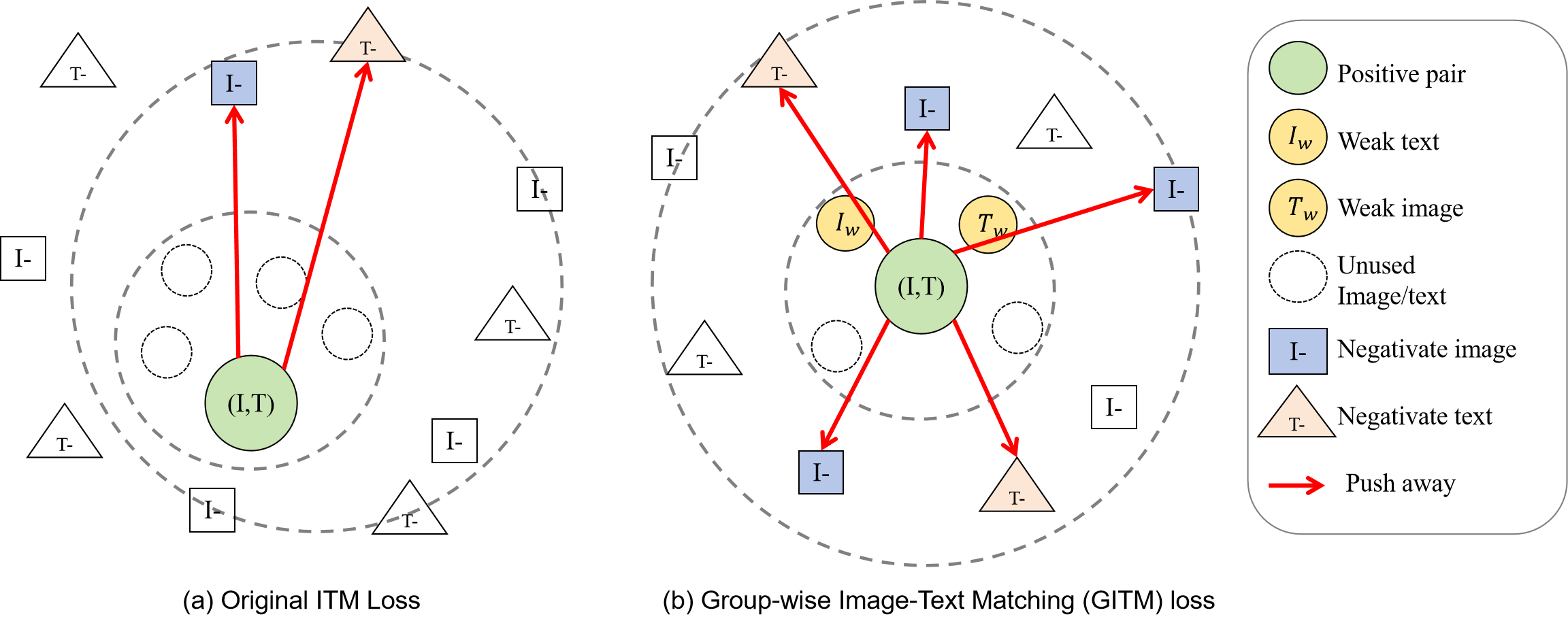}
  \caption{ Intuitive illustration of our Group-wise Image-Text Matching (GITM) loss approach. (a) In the conventional ITM loss calculation, one pair of positive pairs and two negative pairs are used, resulting in limited negative pair diversity. This lack of diversity leads to a skewed representation space distribution, potentially reducing the accuracy of the model performance. Additionally, conventional ITM does not fully leverage all available image-text data, causing semantic deviations between images and texts captured from different perspectives. These deviations can further hinder the ability of the model to effectively learn positive and negative pair matching. (b) Our Group-wise Image-Text Matching (GITM) approach introduces weak positive pairs, allowing the model to learn a more robust latent space for positive pairs while accounting for more diverse scenarios. By utilizing a larger and more diverse set of negative pairs, GITM increases both the number and difficulty of these pairs, resulting in a more evenly distributed representation space and, consequently, enhanced learning accuracy.
  }
  \label{fig:3}
\end{figure}

For each anchor matched pair $(I_i,T_i)$ in a mini-batch, we form a group by additionally sampling one weak-positive image $I^w_i$ and one weak-positive text $T^w_i$ from the same identity, yielding two weak-positive pairs $(I_i,T^w_i)$ and $(I^w_i,T_i)$ (together with the original strong pair $(I_i,T_i)$).
Hard negatives are mined within the current mini-batch based on cosine similarity of the current embeddings $s_{ij}=\cos(f^I_i,f^T_j)$: for each anchor identity, we only consider unpaired samples from different identities ($\mathrm{id}(j)\neq \mathrm{id}(i)$) and select the top-$K$ most similar ones as hard negatives in both image$\rightarrow$text and text$\rightarrow$image directions.
In our implementation, all constructed pairs of a mini-batch are flattened and evaluated by the same ITM classifier, and the loss is computed by averaging over the constructed pairs per anchor group.

To make the above construction explicit, we denote the binary ITM log-likelihood term as $\ell(I,T,p)=p\log\hat{p}(I,T) + (1-p)\log(1-\hat{p}(I,T))$, where $p{=}1$ for matched pairs and $p{=}0$ for negatives.
Let $\mathcal{N}^{T}_i$ be the indices of top-$K$ hard negative texts for $I_i$, and $\mathcal{N}^{I}_i$ be the indices of top-$K$ hard negative images for $T_i$ mined in the batch.
Then the two GITM auxiliary branches in Eqs.~(\ref{eq:5})--(\ref{eq:6}) are instantiated by averaging one weak-positive term and its mined negatives as follows:
\begin{align}
\mathcal{L}_{gitm}^{txt}
= \mathbb{E}_{i}\left[\frac{1}{1+K}\left(\ell(I_i,T^w_i,1) + \sum_{j\in\mathcal{N}^{T}_i}\ell(I_i,T_j,0)\right)\right],
\label{eq:gitm_group_txt}
\\
\mathcal{L}_{gitm}^{img}
= \mathbb{E}_{i}\left[\frac{1}{1+K}\left(\ell(I^w_i,T_i,1) + \sum_{j\in\mathcal{N}^{I}_i}\ell(I_j,T_i,0)\right)\right].
\label{eq:gitm_group_img}
\end{align}

With the standard ITM term on the strong pair $(I_i,T_i)$ providing two directional mined negatives, and the two weak-positive branches each attaching $K$ mined negatives, the overall per-anchor ratio becomes three matched pairs versus $2+2K$ mined negatives (neg3v4 uses $K{=}1$, while neg3v6 uses $K{=}2$).

\noindent\textbf{Implementation summary of GITM group.}
For re-implementation clarity, Table~\ref{tab:gitm_summary} summarizes the per-anchor construction used by GITM. It does not introduce any new component; it only restates the weak-positive sampling, hard-negative mining, and loss evaluation procedure in a compact implementation-oriented form.

\begin{table}[t]
\centering
\footnotesize
\setlength{\tabcolsep}{3pt}
\renewcommand{\arraystretch}{1.03}
\caption{Compact implementation summary of GITM for one anchor pair $(I_i,T_i)$.}
\label{tab:gitm_summary}
\begin{tabularx}{\linewidth}{@{}p{0.19\linewidth}X@{}}
\toprule
\textbf{Step} & \textbf{Summary} \\
\midrule
Anchor pair
& Start from one strong matched image-text pair $(I_i,T_i)$. \\

Weak positives
& Sample one weak-positive image $I_i^{w}$ and one weak-positive text $T_i^{w}$ from the same identity, forming two weak-positive pairs $(I_i,T_i^{w})$ and $(I_i^{w},T_i)$. \\

Hard negatives
& Within the current mini-batch, only consider unpaired samples from different identities. The standard ITM term on the strong pair $(I_i,T_i)$ provides two directional hard negatives (one image$\rightarrow$text and one text$\rightarrow$image). In addition, for the two weak-positive branches, we mine the top-$K$ most similar negatives in each direction based on the current embeddings. \\

Per-anchor group
& The final group contains three matched pairs, i.e., $(I_i,T_i)$, $(I_i,T_i^{w})$, and $(I_i^{w},T_i)$, together with $2+2K$ mined negatives in total. Thus, neg3v4 uses $K=1$, while neg3v6 uses $K=2$. \\

Loss evaluation
& Flatten all constructed pairs in the mini-batch, evaluate them using the same ITM classifier, and average the loss over anchor groups. \\
\bottomrule
\end{tabularx}
\end{table}

To further facilitate the learning from weak positive pairs, we propose Group-wise Image-Text Matching (GITM) loss, shown in Fig.~\ref{fig:3}. Different from the original ITM loss, we include more negative pairs as well as the weak positive pairs. 
To sample the pair, we calculate the similarity between the positive image and the weak positive text, and vice versa. We select multiple hard negative samples and weak positive samples. 
Similarly, the image-text GITM loss $\mathcal{L}_{gitm}^{txt} $ and text-image GITM $\mathcal{L}_{gitm}^{img}$ based on the weak positive pairs can be formulated as: 
\begin{align}
\mathcal{L}_{gitm}^{txt} &= \mathbb{E} \left [ p \log\hat{p} (I,T_{w})   +\left (  1-p \right )  \log\left ( 1- \hat{p} (I,T_{w})\right )\right ], \label{eq:5} \\
\mathcal{L}_{gitm}^{img} &= \mathbb{E} \left [ p \log\hat{p} (I_{w},T)  +\left (  1-p \right )  \log\left ( 1- \hat{p} (I_{w},T)\right )\right ]  , \label{eq:6}
\end{align}
where $p$ is 1 if $(I, T_{w}) $ or $(I_{w}, T) $ is matched, 0 otherwise, and $\hat{p}$ is the match score prediction of image and weak text pairs or weak image and text pairs calculated by an MLPs with Sigmoid activation.
In summary, the final loss of our model can be formulated as the following form:
\begin{align}
   \mathcal{L}_{total} & = \mathcal{L}_{mlm} + \mathcal{L}_{itc} + \mathcal{L}_{itm} + \alpha \mathcal{L}_{uitc} + \beta( \mathcal{L}_{gitm}^{txt} +  \mathcal{L}_{gitm}^{img} ),
   \label{eq:loss}
\end{align}
where $\alpha$ controls the contribution of the uncertainty-aware contrastive term $\mathcal{L}_{uitc}$ in the multi-loss objective, and $\beta$ controls the strength of the auxiliary GITM regularization terms $\mathcal{L}{gitm}^{txt}$ and $\mathcal{L}{gitm}^{img}$ in the multi-loss objective.
We select $\alpha$ via the dedicated sweep in Table~\ref{tab:alpha} under the same setting, and use the best-performing value ($\alpha=0.5$) for all remaining experiments.
For the GITM branch, we keep a small fixed coefficient ($\beta=0.1$) so that $\mathcal{L}_{gitm}^{txt}$ and $\mathcal{L}_{gitm}^{img}$ act as auxiliary stabilizers rather than dominating the optimization when combined with $\mathcal{L}_{itc}$ and $\mathcal{L}_{itm}$.
Importantly, the systematically validated factor for GITM is the \emph{group construction and negative count} (\ie, the positive-to-negative ratio induced by hard-negative mining), which is ablated in Table~\ref{tab:ab1} by comparing neg3v4 and neg3v6.
This confirms that increasing the number of hard negatives per group is the primary driver for the additional gains brought by GITM, while the loss-weight is kept fixed across settings.

\noindent\textbf{Why is $u_w$ used in ITC but not in GITM?}
The two branches serve different purposes in our framework. The uncertainty score $u_w$ is introduced in the ITC branch to regulate the \emph{pair-level ambiguity} of weak positives in the contrastive space: when a weak pair is less reliable, its contribution is softly reduced, whereas more consistent weak pairs provide stronger auxiliary supervision. By contrast, GITM is designed as an auxiliary \emph{binary matching} branch to enrich the \emph{group structure} by explicitly introducing weak positives together with more and harder negatives. Therefore, in GITM, weak positives are assigned the standard matched label, while the key design factor is the group construction / negative count rather than uncertainty reweighting. In this way, uncertainty-aware soft regulation is handled in ITC, whereas group-wise structural enrichment is handled in GITM.

\textbf{Discussion. (1) Why is the proposed ITC loss based on uncertainty adjustment effective?} The proposed method introduces additional uncertainty through weak orthogonal modeling. 
If the uncertainty value is high, it denotes the semantic gap between text and image in the weak positive pairs is large. Therefore, we automatically decrease the loss to mitigate the negative impact. 
If the uncertainty is small, we leverage the weak positive pairs as the positive pairs as auxiliary supervision. 
\textbf{(2) What is the motivation for GITM loss?} Group-wise Image-Text Matching (GITM) enables us to fully leverage the image and text features of weak positive pairs (as shown in Fig.~\ref{fig:3}). This strategy allows the model to extract a more comprehensive relationship between multiple pairs. In experiment, we increase the number of positive and negative pairs, altering the ratio from 1:2 to 3:6 (comprising one pair of strong positive samples and two pairs of weak positive samples, and six negative pairs). 
We observe that larger group-wise metric learning boosts the diversity of negative pairs and thus, enhances the model ability to discriminate between more negatives.

\section{Experiment} \label{experiment}

\subsection{Datasets and Evaluation Protocol}~\label{datasets}
\textbf{Datasets}. We employ the synthetic dataset MALS~\cite{APTM} for pre-training, which comprises 1,510,330 image-text pairs, each annotated with relevant attribute labels. We validate our method on three benchmark datasets.
For fine-tuning and evaluation, we utilize widely-used datasets: CUHK-PEDES~\cite{Li_Xiao_Li_Zhou_Yue_Wang_2017}, RSTPReid~\cite{zhu2021dssl}, and ICFG-PEDES~\cite{Ding_Ding_Shao_Tao_2021}. CUHK-PEDES integrates 40,206 images of 13,003 individuals from five person search datasets: CUHK03~\cite{Li_Zhao_Xiao_Wang_2014}, Market-1501~\cite{Zheng_Shen_Tian_Wang_Bu_Tian_2015}, SSM~\cite{Xiao_Li_Wang_Li_Wang_2016}, VIPER~\cite{Gray_Brennan_Tao_2007}, and CUHK01~\cite{Li_Zhao_Wang_2013}. Each image is annotated with two sentences, totaling 80,412 descriptions. RSTPReid includes 20,505 images of 4,101 individuals and is constructed from MSMT17~\cite{Wei_Zhang_Gao_Tian_2018}. Each identity has five images captured by different cameras, with each image paired with two textual descriptions. ICFG-PEDES, also derived from MSMT17, consists of 54,522 images of 4,102 individuals, each accompanied by one textual description. Our method is evaluated on the three public text-based person search datasets: CUHK-PEDES, RSTPReid, and ICFG-PEDES.

\textbf{Evaluation metrics}. 
Following previous works on text-based person search, we adopt the mean Average Precision (AP) and Recall@1,5,10 as our primary evaluation metrics. The Recall@K, whose value is 1 if the first matched image has appeared before the K-th image. Recall@K is sensitive to the position of the first matched image and suits the test set with only one true-matched image in the gallery. The average precision (AP) is the area under the PR (Precision-Recall) curve,  considering all ground-truth images in the gallery. mAP is calculated and averaged for the average accuracy (AP) of each category.

\textbf{Implementation Details}~\label{details}
Our model is based on the current advanced two-stage benchmark model, and all experiments are trained using Pytorch on eight NVIDIA A800 GPUs. In pre-training, the Model image encoder uses Swinv2-B as the backbone model~\cite{SwinTransformer}. Text encoder and cross encoder use BERT-base~\cite{devlin2019bert}, respectively. The first 6 and last 6 layers are initialized. At the same time, for the pre-training dataset MALS, we adopt the data filtering method~\cite{oursMM} to screen and retrain the image text dataset with a high matching degree for pre-training. We pre-train the model on 32 epochs with a small batch size of 70 per GPU. We use the AdamW~\cite{Loshchilov_Hutter_2017} optimizer with a weight attenuation of 0.01. In the first 2600 steps learning rate from $1e^{-5}$ begins to warm up, according to the linear plan, and then from $1e^{-4}$ goes down to $1e^{-5}$. Each image input is adjusted to 384 $\times$ 384. Random horizontal inversion, RandAugment~\cite{Cubuk_Zoph_Shlens_Le_2020}, and random erase~\cite{Zhong_Zheng_Kang_Li_Yang_2020} are used for image enhancement. In addition to the image data enhancement mentioned in the pre-training, we also adopt EDA~\cite{Wei_Zou_2019} for text data enhancement and set the small batch size to 35. After pre-training, the model is fine-tuned for 30 epochs on the downstream dataset, with a small batch size of 35 per GPU. Set the learning rate to $ 1e^{-4} $ in the image Encoder, and warm up for the first 3 epochs. Then a linear scheduler is applied to gradually attenuate the learning rate. In the finetune stage, different images and texts with the same ID are randomly selected as weak pairs for training.
At the same time, for the MALS dataset used in the pre-training stage, we implement a data filtering strategy~\cite{oursMM} to remove the low-quality training data.

\noindent\textbf{Compute/Memory Overhead.}
Our proposed GITM does not introduce any additional learnable modules or parameters; it only modifies the ITM pair construction by incorporating weak positives and group-wise hard negatives.
Under the same training setup described above (same backbone/model configuration, batch size, and optimization settings), GITM (neg3v6) increases the wall-clock training time per epoch from 9:47 to 13:50 and the peak GPU memory from 70776\,MiB to 80060\,MiB.
This overhead is expected since neg3v6 expands the ITM supervision to a group-wise composition with 3 positive pairs (1 strong + 2 weak) and 6 negative pairs, thus evaluating more image--text pairs within the ITM branch.
We include this to make the efficiency trade-off of the proposed pairing strategy explicit. \noindent\textbf{Optimization of $\gamma$.}
To ensure numerical stability and enforce the positivity constraint, we parameterize the learnable scale $\gamma$ in log-space and recover it via an exponential mapping. In practice, we optimize a scalar parameter (\texttt{log\_gamma}) initialized by $\log(1.0)$ and compute $\gamma=\exp(\texttt{log\_gamma})$ during training. This guarantees $\gamma>0$ throughout optimization and avoids non-positive scaling.
Importantly, this computational overhead is strictly limited to the training phase. At inference, the retrieval speed remains unchanged from the baseline, since the proposed method introduces no additional learnable modules and leaves the evaluation pipeline unchanged. Under the same single-GPU A800 evaluation setting, both the baseline and our method showed an inference time of about 9 minutes, confirming that the proposed training-time modifications do not introduce additional test-time latency.

\begin{table}[t!]
\centering
\vspace{-3em}
\setlength{\tabcolsep}{20pt}
\renewcommand{\arraystretch}{0.9}
\caption{Performance comparison on CUHK-PEDES. Here we show the performance of the previous methods on the recall@1,5,10, mAP in \%. For a fair comparison, we change the backbone of the baseline. Baseline: We re-implement APTM~\cite{APTM} with backbone Swinv2-B. $^{*}$ indicates the use of additional information, \eg, human parsing. }
\scalebox{0.8}{\small
\begin{tabular}{l|cccc}
\toprule
\multicolumn{1}{c|}{Method}                                                                          & R@1   & R@5   & R@10  & mAP   \\ \midrule
Dual Path~\cite{Zheng_Zheng_Garrett_Yang_Xu_Shen_2020}  & 44.40 & 66.26 & 75.07 & -     \\
MIA~\cite{niu2019improving}                                  & 53.10 & 75.00 & 82.90 & -     \\
DSSL~\cite{zhu2021dssl}                                     & 59.98 & 80.41 & 87.56 & -     \\
SSAN~\cite{Ding_Ding_Shao_Tao_2021}                      & 61.37 & 80.15 & 86.73 & -     \\
TIPCB~\cite{Chen_Zhang_Lu_Wang_Zheng_2022}              & 63.63 & 82.82 & 89.01 & -     \\
LBUL~\cite{Wang_Zhu_Xue_Wan_Liu_Wang_Li_2022}         & 64.04 & 82.66 & 87.22 & -     \\
CAIBC~\cite{CAIBC}                                           & 64.43 & 82.87 & 88.37 & -     \\
LGUR~\cite{Shao_Zhang_Fang_Lin_Wang_Ding_2022}         & 65.25 & 83.12 & 89.00 & -     \\
TransTPS~\cite{TransTPS}         & 68.23 & 86.37 & 91.65 & -     \\
CFine~\cite{yan2022clipdriven}                               & 69.57 & 85.93 & 91.15 & -     \\
VGSG~\cite{10345496}                                          & 71.38 & 86.75 & 91.86 & 67.91 \\
MACF~\cite{Sun2024AnAC}                                             & 73.33 &88.57 &93.02  & - \\
IRRA~\cite{jiang2023crossmodal}                              & 73.38 & 89.93 & 93.71 & 66.13 \\
TBPS-CLIP~\cite{cao2023empirical}                            & 73.54 & 88.19 & 92.35 & 65.38 \\
SAMC~\cite{samc}                         & 74.03 & 89.18 & 93.31 & 68.42 \\
RDE~\cite{RDE}                         & 75.94 & 90.14 & 94.12 & 67.56 \\
RaSa~\cite{Bai_2023}                                        & 76.51 & 90.29 & 94.25 & 69.38 \\
APTM~\cite{APTM}                                             & 76.53 & 90.04 & 94.15 & 66.91 \\ 
ITSELF~\cite{nguyen2026itself}   & 76.95 & 90.64 & 94.36 & 69.38 \\
DiCo~\cite{KIM2026132885}                         & 77.21 & \textbf{91.85} & \textbf{95.63} & - \\
AUL~\cite{AUL}                         & 77.23 & 90.43 & 94.41 & - \\
BAMG$^{*}$~\cite{BAMG}                         & 79.98 & 92.31 & 94.03 & 68.55 \\
\midrule
Baseline                                                                     & 76.90 & 90.75 & 94.33 & 68.85 \\
Ours                                                                          & \textbf{77.88} & 91.05 & 94.57 & \textbf{72.44} (\textcolor{red}{+3.59}) \\ \bottomrule
\end{tabular}
 }
\label{tab:cuhk}
\end{table}


\subsection{Comparison with State-of-the-art Methods}~\label{comparsions}
On three benchmark datasets, CUHK-PEDES, RSTPReid, and ICFG-PEDES, we compare the proposed method with other advanced text-based person retrieval methods that have reported results or can be re-implemented. The performance evaluation indicators are mean Average Precision (AP) and Recall@1,5,10. 

\textbf{Performance comparison on CUHK-PEDES:}
We compare our method with lots of competitive methods on CUHK-PEDES. The performance of our method on CUHK-PEDES is shown in Table~\ref{tab:cuhk}, from which we can observe that:
\begin{itemize}[leftmargin=*,labelsep=0.8em]
\item(1) Our method achieves a state-of-the-art mAP of 72.44\%, along with leading performance in Recall@1, Recall@5, and Recall@10, with scores of 77.88\%, 91.05\%, and 94.57\%, respectively, significantly outperforming other methods. In particular, in terms of mAP, the accuracy of models with uncertainty is improved by +3.06\% over RaSa~\cite{Bai_2023} on CUHK-PEDES.
\item(2) Comparing to our re-implemented baseline, \ie, APTM + Swinv2-B (Recall@1/5/10: 76.90\%, 90.75\%, 94.33\%, mAP: 68.85\%), which adopts conventional contrastive learning and image-text matching objectives, our proposed uncertainty-aware method achieves a +3.59\% improvement in mAP and a +0.98\% improvement in Recall@1. 
Notably, the much larger gain in mAP than in Recall@K suggests improved ranking quality beyond top-K hits, \ie, more positive samples are promoted to higher positions throughout the ranked retrieval list, which is better reflected by mAP.

\item(3) At the same time, we can observe that the proposed method outperforms the source domain model, \ie, APTM (Recall@1, 5, 10: 76.53\%, 90.04\%, 94.15\%, mAP: 66.91\%). This indicates that our uncertainty-based approach effectively leverages a broader range of sample information, leading to a more balanced representation space distribution. By employing group-wise ITM, the model is exposed to a greater diversity of negative samples, significantly contributing to the overall performance improvement. 

\item(4) The proposed method also surpasses, \ie, RaSa (mAP: 69.38\%), which employs relation and sensitivity-aware representation learning. Our uncertainty-aware approach proves more effective in achieving mAP improvements. By employing our proposed uncertainty-based approach, the model can more effectively utilize weak positive pairs to learn a richer feature representation space. This enhancement facilitate the model to learn discriminative feature, allowing the model to correctly identify and rank more positive candidates at higher retrieval hierarchy.

\end{itemize}

\begin{table*}[t!]
\centering

\begin{minipage}[t]{0.49\textwidth}
\centering
\vspace{-2em}
\setlength{\tabcolsep}{3pt}
\renewcommand{\arraystretch}{1.1}
\captionof{table}{Performance comparison on RSTPReid. 
Baseline: We re-implement APTM~\cite{APTM} with backbone Swinv2-B. $^{*}$ indicates the use of additional information (humman parsing). }
\scalebox{0.8}{\small
\begin{tabular}{l|cccc}
\toprule
\multicolumn{1}{c|}{Method}                                                                       & R@1   & R@5   & R@10  & mAP   \\ \midrule
DSSL~\cite{zhu2021dssl}                                    & 32.43 & 55.08 & 63.19 & -     \\
LBUL~\cite{Wang_Zhu_Xue_Wan_Liu_Wang_Li_2022}       & 45.55 & 68.20 & 77.85 & -     \\
IVT~\cite{Shu_Wen_Wu_Chen_Song_Qiao_Ren_Wang_2022} & 46.70 & 70.00 & 78.80 & -     \\
CAIBC~\cite{CAIBC}                                         & 47.35 & 69.55 & 79.00 & -     \\
CFine~\cite{yan2022clipdriven}                             & 50.55 & 72.50 & 81.60 & -     \\
TransTPS~\cite{TransTPS}         & 56.05 & 78.65 & 86.75 & -     \\
IRRA~\cite{jiang2023crossmodal}                            & 60.20 & 81.30 & 88.20 & 47.17 \\
SAMC~\cite{samc}  & 60.80 & 82.35 & 89.00 & 49.67 \\
TBPS-CLIP~\cite{cao2023empirical}                          & 61.95 & 83.55 & 88.75 & 48.26 \\
RaSa~\cite{Bai_2023}                                     & 66.90 & 86.50 & 91.35 & 52.31 \\
APTM~\cite{APTM}                                            & 67.50 & 85.70 & 91.45 & 52.56 \\
RDE~\cite{RDE}  & 65.35 & 83.95 & 89.90 & 50.88 \\
ITSELF~\cite{nguyen2026itself}   & 67.30 & 85.60  & 90.50 & 53.05 \\
DiCo~\cite{KIM2026132885}                         & 67.84 & 85.72 & 91.98 & - \\
BAMG$^{*}$~\cite{BAMG}                         & 69.73 &  87.65 & 93.33 & 55.21 \\
AUL~\cite{AUL}    & \textbf{71.65} & \textbf{87.55} & \textbf{92.05} & - \\
 \midrule
Baseline                                                                   & 66.75 & 85.70 & 91.65 & 53.22 \\
Ours                                                                        & 69.45 & 85.50 & 91.65 & \textbf{56.11} (\textcolor{red}{+2.89}) \\ \bottomrule
\end{tabular}
}
\label{tab:RSTP}
\end{minipage}
\hfill
\begin{minipage}[t]{0.49\textwidth}
\centering
\vspace{-2em}
\setlength{\tabcolsep}{3pt}
\renewcommand{\arraystretch}{1.1}
\captionof{table}{Performance comparison on ICFG-PEDES. 
Baseline: We re-implement APTM~\cite{APTM} with backbone Swinv2-B. $^{*}$ indicates the use of additional information, \eg, humman parsing. }
\scalebox{0.8}{\small
\begin{tabular}{l|cccc}
\toprule
\multicolumn{1}{c|}{Method}      & R@1    & R@5    & R@10   & mAP    \\ \midrule
Dual Path~\cite{Zheng_Zheng_Garrett_Yang_Xu_Shen_2020}  & 38.99  & 59.44  & 68.41  & -     \\
MIA~\cite{niu2019improving}       & 46.49  & 67.14  & 75.18  & -     \\
SCAN~\cite{lee2018stacked}     & 50.05  & 69.65  & 77.21  & -     \\
SSAN~\cite{Ding_Ding_Shao_Tao_2021}     & 54.23  & 72.63  & 79.53  & -      \\
IVT~\cite{Shu_Wen_Wu_Chen_Song_Qiao_Ren_Wang_2022}      & 56.04  & 73.60  & 80.22  & -      \\
LGUR~\cite{Shao_Zhang_Fang_Lin_Wang_Ding_2022}     & 59.02  & 75.32  & 81.56  & -      \\
CFine~\cite{yan2022clipdriven}     & 60.83  & 76.55  & 82.42  & -      \\
MACF~\cite{Sun2024AnAC}       & 62.95 &79.93 &85.04  & - \\
IRRA~\cite{jiang2023crossmodal}     & 63.46  & 80.25  & 85.82  & 38.06   \\
SAMC~\cite{samc}                         & 63.68 & 79.69 & 85.21 & 42.41 \\
TBPS-CLIP~\cite{cao2023empirical}  &65.05  &80.34	&85.47	&39.83  \\
RaSa~\cite{Bai_2023}     & 65.28  & 80.04  & 85.12  & 41.29  \\
RDE~\cite{RDE}                         & 67.68 & 82.47 & 87.36 & 40.06 \\
APTM~\cite{APTM}   & 68.51  & 82.99  & 87.56  & 41.22  \\
DiCo~\cite{KIM2026132885}    & 67.81 & 83.29 & 87.62 & - \\
AUL~\cite{AUL}    & 69.16 & 83.32 & 88.37 & - \\
ITSELF~\cite{nguyen2026itself}   & \textbf{69.23} & 82.84 & 87.62 & 43.80 \\
BAMG$^{*}$~\cite{BAMG}                         & 71.70 & 86.34 & 89.71 & 42.37 \\
\midrule
Baseline    & 68.71  & \textbf{83.67}  & \textbf{88.39}  & 44.28  \\
Ours       & 69.22 & 83.56 & 88.13 & \textbf{48.23} (\textcolor{red}{+3.95})  \\ 
\bottomrule
\end{tabular}
}
\label{tab:icfg}
\end{minipage}

\end{table*}

\textbf{Performance comparison on RSTPReid and ICFG-PEDES:}
The performance of our model on RSTPReid and ICFG-PEDES is shown in Table~\ref{tab:RSTP} and Table~\ref{tab:icfg} respectively, and we can observe similar performance improvement: (1) The proposed method is significantly superior to other models, obtaining 69.45\% Recall@1, 85.50\% Recall@5, 91.65\% Recall@10 and 56.11\% mAP on RSTPReid. On ICFG-PEDES, 69.22\% Recall@1, 83.56\% Recall@5, 88.13\% Recall@10 and 48.23\% of mAP are obtained. In particular, in terms of mAP, the accuracy of models with uncertainty is improved by +3.55\% over APTM~\cite{APTM} on RSTPReid and +6.94\% over RaSa~\cite{Bai_2023} on ICFG-PEDES. (2) Using the same Swinv2-B backbone, the proposed method achieves competitive results on RSTPReid and ICFG-PEDES, with mAP improvements of +2.89\% and +3.95\%, respectively. Additionally, We achieve significant improvements of +2.70\% and +0.51\% in Recall@1. Our proposed uncertainty-aware ITC and group-wise ITM approach enhances model retrieval capabilities across both datasets by leveraging diverse weak positive samples as a supplement and incorporating more negative samples into ITM learning through a group-wise method. This approach enables the model to retrieve more correctly ranked positive samples, resulting in a significant improvement in mAP performance.

\begin{table*}[t!]
\centering
\setlength{\tabcolsep}{10pt}
\renewcommand{\arraystretch}{1.1}
\caption{Comparison results (\%) on the domain generalization tasks (\ie, CUHK-PEDES to ICFG-PEDES (C $\rightarrow$ I) and ICFG-PEDES to CUHK-PEDES (I $\rightarrow$ C)). The \textbf{bold} and \underline{underline} texts denote the best and runner-up results, respectively.}
\small
\scalebox{0.8}{
\begin{tabular}{l|c c c|c c c}
\toprule
\multirow{2}{*}{Method} & \multicolumn{3}{c|}{I $\rightarrow$ C} & \multicolumn{3}{c}{C $\rightarrow$ I} \\
\cmidrule(lr){2-4}\cmidrule(lr){5-7}
& R@1 & R@5 & R@10 & R@1 & R@5 & R@10 \\
\midrule
Dual Path~\cite{Zheng_Zheng_Garrett_Yang_Xu_Shen_2020} & 15.41 & 29.80 & 38.19 & 7.63  & 17.14 & 23.52 \\
MIA~\cite{niu2019improving}                           & 19.35 & 36.78 & 46.42 & 10.93 & 23.77 & 32.39 \\
SCAN~\cite{lee2018stacked}                            & 21.27 & 39.26 & 48.83 & 13.63 & 28.61 & 37.05 \\
SSAN~\cite{Ding_Ding_Shao_Tao_2021}                   & 24.72 & 43.43 & 53.01 & 16.68 & 33.84 & 43.00 \\
LGUR~\cite{Shao_Zhang_Fang_Lin_Wang_Ding_2022}        & 34.25 & 52.58 & 60.85 & 25.44 & 44.48 & 54.39 \\
VGSG~\cite{10345496}                                  & \underline{35.85} & \underline{55.04} & \underline{63.61} 
                                                     & \underline{27.17} & \underline{47.77} & \underline{57.27} \\
\midrule
\textbf{Ours}                                        
& \textbf{47.19} & \textbf{70.27} & \textbf{78.09}
& \textbf{49.33} & \textbf{68.61} & \textbf{75.79} \\
\bottomrule
\end{tabular}}
\label{tab:DG1}
\end{table*}

\textbf{Performance comparison on the Domain Generalization (DG) task.}
Our method effectively utilizes information from weak positive image-text pairs as supplementary. This approach promotes a more uniform distribution in the model representation space, which naturally suggests that the model can generalize well to other domains. To validate this, we conduct experiments on Domain Generalization (DG) tasks. Specifically, we directly deploy the model, pre-trained on the source domain, to target datasets without further fine-tuning.
As shown in Table~\ref{tab:DG1}, our method outperforms all other compared approaches. Notably, our method surpasses VGSG~\cite{10345496} by +11.34\% in Rank-1 accuracy on the C $\rightarrow$ I task, and by +22.16\% in Rank-1 accuracy on the I $\rightarrow$ C task.
In Table~\ref{tab:DG2}, we present the performance of our method on four additional Domain Generalization (DG) tasks. These experiments demonstrate that our uncertainty-aware method exhibits strong generalization capabilities.

To further isolate the contribution of each proposed component under domain shift, we additionally conduct a cross-domain ablation study under the same source-only transfer protocol on the representative I$\rightarrow$C and C$\rightarrow$I tasks. As shown in Table~\ref{tab:DG_ablation}, both uncertainty-aware ITC and GITM remain effective in the cross-domain scenario, and their combination yields the strongest overall transfer performance. On I$\rightarrow$C, the full model improves R@1/mAP from 45.96/40.85 to 47.19/44.10. On C$\rightarrow$I, it improves R@1/mAP from 47.06/25.60 to 49.33/28.53. These results further support that the DG gains stem from the proposed methodology itself rather than target-side adaptation, since all evaluations are conducted without target-domain fine-tuning.

\begin{table*}[t]
\centering
\setlength{\tabcolsep}{15pt}
\renewcommand{\arraystretch}{1.1}
\caption{Our results (\%) on the domain generalization tasks (\ie, CUHK-PEDES to RSTPReid (C $\rightarrow$ R) and RSTPReid to CUHK-PEDES (R $\rightarrow$ C), RSTPReid to ICFG-PEDES (R $\rightarrow$ I) and ICFG-PEDES to RSTPReid (I $\rightarrow$ R)). }
\small
\scalebox{0.9}{
    \centering
    \begin{tabular}{c|l|c c c}
        \toprule
        Tasks & \multicolumn{1}{c|}{Method} & R@1 & R@5 & R@10 \\
        \midrule
        \midrule
        \multirow{1}{*}{\shortstack{C  $\rightarrow$  R}} 
        & Ours & 56.35 & 77.30 & 85.30 \\

        \multirow{1}{*}{\shortstack{R $\rightarrow$  C}} 
        & Ours & 39.49 & 61.88 & 70.63 \\
        \midrule
        \midrule
        \multirow{1}{*}{\shortstack{R  $\rightarrow$  I}} 
        & Ours & 45.04 & 60.71 & 67.28 \\

        \multirow{1}{*}{\shortstack{I $\rightarrow$  R}} 
        & Ours & 55.70 & 74.55 &  82.45 \\
        \bottomrule
    \end{tabular}
    }
    \label{tab:DG2}
\end{table*}
\noindent\textbf{Discussion.}
With the added comparisons to recent CLIP-based TBPS systems (e.g., ITSELF, IRRA/RDE, BAMG, DiCo), our method achieves the best mAP performance on CUHK-PEDES and RSTPReid, and ICFG-PEDES while providing a consistent gain over the strong re-implemented baseline across all benchmarks.
Notably, our contribution is \emph{pair-level} (uncertainty-aware optimization for noisy/ambiguous correspondences) and is therefore orthogonal to architecture-centric designs (\eg, graph modeling, slot-based disentanglement, fine-grained alignment modules), suggesting potential complementarity when combined.

\begin{table*}[t!]
\centering
\setlength{\tabcolsep}{4pt}
\renewcommand{\arraystretch}{1.2}
\caption{Ablation study of our method with different settings on CUHK-PEDES. The difference between Baseline and Baseline$^\dagger$ is that we re-implement APTM~\cite{APTM} with backbone Swinv2-B for a fair comparison. $\mathcal{L}_{uitc}$ is the optimization objective that uses uncertainty-aware ITC to leverage information about the weak positive pairs fully. $\mathcal{L}_{gitm}$ (neg3v4) is that we adopt the weak positive pairs to increase the number and difficulty of negative pairs by using Group-wise Image-Text Matching (GITM), comprising 1 positive pair, 2 weak positive pairs, and 4 negative pairs. Similarly, $\mathcal{L}_{gitm}$ (neg3v6) is expanded to 1 positive pair, 2 weak positive pairs, and 6 negative pairs.}
\scalebox{0.95}{\small
\begin{tabular}{l|c|c|c|cccc}
\toprule
Method             &  $\mathcal{L}_{uitc}$                & $\mathcal{L}_{gitm}$ (neg3v4)           & $\mathcal{L}_{gitm}$ (neg3v6)             & R@1            & R@5            & R@10           & mAP            \\ \midrule
Baseline    &                           &                           &                           & 76.53          & 90.04          & 94.15          & 66.91          \\
Baseline$^\dagger$ &                           &                           &                           & 76.90          & 90.76          & 94.33          & 68.86          \\
Baseline          & \checkmark &                           &                           & 76.88          & 90.60          & 94.35          & 70.49          \\
Baseline          & \checkmark & \checkmark &                           & 76.85          & 90.77          & 94.54          & 71.89          \\
Baseline          & \checkmark &                           & \checkmark & \textbf{77.88} & \textbf{91.05} & \textbf{94.57} & \textbf{72.44} \\ \bottomrule
\end{tabular}}
\label{tab:ab1}
\end{table*}

\begin{table*}[t!]
\centering
\setlength{\tabcolsep}{10pt}
\renewcommand{\arraystretch}{1.1}
\caption{Cross-domain ablation results (\%) under the source-only transfer protocol on the domain generalization tasks, i.e., ICFG-PEDES $\rightarrow$ CUHK-PEDES (I $\rightarrow$ C) and CUHK-PEDES $\rightarrow$ ICFG-PEDES (C $\rightarrow$ I). ``Baseline$^\dagger$ + $\mathcal{L}_{uitc}$'' uses only the uncertainty-aware ITC objective, ``Baseline$^\dagger$ + $\mathcal{L}_{gitm}$'' uses only the GITM objective, and ``Ours'' combines both components. All models are trained on the source domain and directly evaluated on the target domain without target-domain fine-tuning. \textbf{Bold} denotes the best result in each column.}
\small
\scalebox{0.75}{
\begin{tabular}{l|c c c c|c c c c}
\toprule
\multirow{2}{*}{Method} & \multicolumn{4}{c|}{I $\rightarrow$ C} & \multicolumn{4}{c}{C $\rightarrow$ I} \\
\cmidrule(lr){2-5}\cmidrule(lr){6-9}
& R@1 & R@5 & R@10 & mAP & R@1 & R@5 & R@10 & mAP \\
\midrule
Baseline$^\dagger$ 
& 45.96 & 68.52 & 76.51 & 40.85
& 47.06 & 68.13 & 75.02 & 25.60 \\

Baseline$^\dagger$ + $\mathcal{L}_{uitc}$ 
& 45.87 & 68.67 & 76.43 & 41.81
& 47.66 & 68.35 & 75.15 & 26.39 \\

Baseline$^\dagger$ + $\mathcal{L}_{gitm}$ (neg3v6)  
& 45.35 & 69.33 & 77.62 & 41.88
& 48.44 & \textbf{68.83} & \textbf{75.91} & 27.21 \\

\textbf{Ours}                                        
& \textbf{47.19} & \textbf{70.27} & \textbf{78.09} & \textbf{44.10}
& \textbf{49.33} & 68.61 & 75.79 & \textbf{28.53} \\
\bottomrule
\end{tabular}}
\label{tab:DG_ablation}
\end{table*}

\subsection{Ablation Studies and Further Discussion}~\label{ablations}
To further evaluate our approach, we conduct several ablation studies, with a primary focus on the fine-tuning stage. This emphasis is because our methodology aims to enhance model performance through targeted adjustments to model loss during the fine-tuning phase.
\par
\textbf{Effect of our uncertainty-aware ITC loss and group-wise ITM loss.} 
We show the ablation comparison of our completely proposed experimental methods in Table~\ref{tab:ab1}. (1) First, we filter the dataset based on \cite{oursMM} and conduct experiments based on the initial baseline~\cite{APTM}. (2) Second, the image encoder backbone is replaced by Swinv2-B (the input image size is adjusted to 384 $\times$ 384), which shows that the learning ability of the model is further improved. (3) Third, we start to replace the backbone model as the baseline and gradually increase our uncertainty method on it. Firstly, we verify the method of applying uncertainty to adjust ITC loss. It can be seen that the model further increases mAP +1.63\% while holding Recall@k. (4) Finally, uncertainty is applied to expand the hard negative in ITM and increase the difficulty of the hard negative. It can be seen that we finally expand to 3 positive pairs and 6 negative pairs. This strategy yields significant improvements in both Recall@k and mAP. Specifically, our method surpasses the baseline (Swinv2-B) on mAP by +3.58\% and on Recall@1 by +0.98\%.

\textbf{Comparison of the hard negative number for group-wise image-text matching (GITM).}
We further evaluate the effect of uncertainty-adjusted hard negative counts in Table~\ref{tab:ab1}. $\mathcal{L}{gitm}$ (neg3v4) expands each example to 3 positive and 4 negative pairs via uncertainty. The set includes the original one-to-one positive pair; two hard negatives obtained by comparing features of the positive pair and selecting the most similar negatives for both image and text; and a new positive formed by pairing the positive image with a weak positive text, with one corresponding negative built from that weak text. Analogously, another positive and its negative are created using a weak positive image. Similarly, $\mathcal{L}{gitm}$ (neg3v6) constructs 3 positive and 6 negative pairs (details in Section~\ref{sec:3.4}). We observe that increasing the number of negatives in GITM is more critical. Adding GITM improves performance over not using it: mAP +0.55\%, Recall@1 +1.03\%, Recall@5 +0.28\%, Recall@10 +0.03\%.

\textbf{Analyze the influence of image feature dimension embedding dimension.}
In Table~\ref{tab:ab2}, we explore the influence of the image feature embedding dimension. 
We observe that increasing the embed dimension from 256 to 2048 leads to improvements across various evaluation metrics, suggesting that a higher embed dimension enhances the model perceptual and learning capabilities. Consequently, all subsequent experiments are conducted with a 2048 embed dimension.
\begin{table*}[t!]
\centering

\begin{minipage}[t]{0.49\textwidth}
\centering
\setlength{\tabcolsep}{7pt}
\renewcommand{\arraystretch}{1.2}
\captionof{table}{Impact of different ITC embedding dimensions on our model. We report the recall rate and mAP on CUHK-PEDES. Here we only deploy the uncertainty-aware ITC loss. We have achieved the best mAP when the embedding dimension is 2048. 
}
\scalebox{0.8}{\small
\begin{tabular}{l|cccc}
\toprule
Embedding\_dim      & \multicolumn{1}{c}{R@1} & \multicolumn{1}{c}{R@5} & \multicolumn{1}{c}{R@10} & \multicolumn{1}{c}{mAP} \\ \midrule
256  & 76.25                   & 90.10                   & 93.84                    & 68.94                   \\
1024 & 76.48                   & 90.16                   & 93.94                    & 68.83                   \\
2048 & 76.25                   & 90.17                   & 93.91                    & 68.96                   \\ \bottomrule
\end{tabular}
}
\label{tab:ab2}
\end{minipage}
\hfill
\begin{minipage}[t]{0.49\textwidth}
\centering
\setlength{\tabcolsep}{10pt}
\renewcommand{\arraystretch}{1.2}
\captionof{table}{Impact of different loss weights $\alpha$ of uncertainty-aware ITC on our model. Here we report the recall rate and mAP on CUHK-PEDES. 
We have achieved the best mAP when the loss weight $\alpha$ of uncertainty-aware ITC is 0.5.} 
\scalebox{0.8}{\small
\begin{tabular}{c|cccc}
\toprule
$\alpha$ & R@1            & R@5            & R@10           & mAP            \\ \midrule
0.3      & 76.58          & \textbf{90.34} & \textbf{94.23} & 70.17          \\
0.4      & 76.53          & 89.96          & 94.07          & 70.22          \\
0.5      & \textbf{76.92} & 90.11          & 94.09          & \textbf{70.78} \\
0.6      & 76.59          & 90.16          & 94.04          & 70.73          \\
0.7      & 76.24          & 90.15          & 94.10          & 70.26          \\ \bottomrule
\end{tabular}
}
\label{tab:alpha}
\end{minipage}

\end{table*}

\textbf{Analyze the influence of loss weight in front of uncertainty-aware ITC loss.}
We study the impact of the loss weight in front of uncertainty-aware ITC loss. In particular,  we change the alpha scale in Eq.~\ref{eq:loss}. We show the effect of different alpha scales on model performance in Table~\ref{tab:alpha}. When comparing the impact of alpha on the model, Swin-B is used as the backbone. In order to verify the impact of alpha on uncertainty-aware ITC, we only apply uncertainty learning to regulate ITC loss. Without using uncertainty to increase hard negative, we can observe that the model gets the best performance with uncertainty when $\alpha=0.5$. At the same time, the experimental results also show that the smaller the proportion of ITC loss adjusted by our uncertainty method in the total loss, the higher the performance of Recall@5, 10, the larger the proportion, and the better the values of mAP and Recall@1, but the higher the proportion is not the better. The experiment shows that the model achieves the best performance when alpha is 0.5.
\begin{table*}[t!]
\centering

\begin{minipage}[t]{0.49\textwidth}
\centering
\setlength{\tabcolsep}{8pt}
\renewcommand{\arraystretch}{1.2}
\caption{Ablation study on the loss weight $\beta$. All experiments are conducted on CUHK-PEDES with a fixed $\alpha=0.1$. Bold denotes the best performance.} 
\scalebox{0.8}{\small
\begin{tabular}{c|cccc}
\toprule
$\beta$ & R@1            & R@5            & R@10           & mAP            \\ \midrule
0.01      & 77.37          & 90.95          & \textbf{94.74}          & 70.23          \\
0.05      & 77.57          & \textbf{91.16}          & 94.41          & 71.68          \\
0.1      & \textbf{77.88}   & 91.05     & 94.57 & 72.44          \\
0.2      & 76.93          & 90.04          & 93.63          & \textbf{72.47}          \\
0.3      & 76.30          & 89.18          & 92.77          & 72.09 \\
0.4      & 76.07          & 88.97          & 92.49          & 72.05          \\
0.5      & 76.18          & 88.56          & 92.28          & 72.06          \\ \bottomrule
\end{tabular}
}
\label{tab:beta}
\end{minipage}
\hfill
\begin{minipage}[t]{0.49\textwidth}
\centering
\setlength{\tabcolsep}{10pt}
\renewcommand{\arraystretch}{1.2}
\caption{Impact of different input image sizes. Here we report the recall rate and mAP on CUHK-PEDES. 
The best mAP is achieved when the input image size is 384 $\times$ 384.
}
\scalebox{0.8}{\small
\begin{tabular}{c|cccc}
\toprule
H $\times$ W     & R@1            & R@5            & R@10           & mAP            \\ \midrule
256 $\times$ 256 & 76.79          & 90.68          & 94.23          & 71.29          \\
384 $\times$ 192 & 77.01          & 90.77          & 94.25          & 70.97          \\
384 $\times$ 384 & \textbf{77.88} & \textbf{91.05} & \textbf{94.57} & \textbf{72.44} \\
576 $\times$ 192 & 76.79          & 90.71          & 94.41 & 71.07          \\ \bottomrule
\end{tabular}
}
\label{tab:ab3}
\end{minipage}

\end{table*}

\begin{figure}[t!]
  \centering
  \vspace{-4em}
  \includegraphics[width=\linewidth]{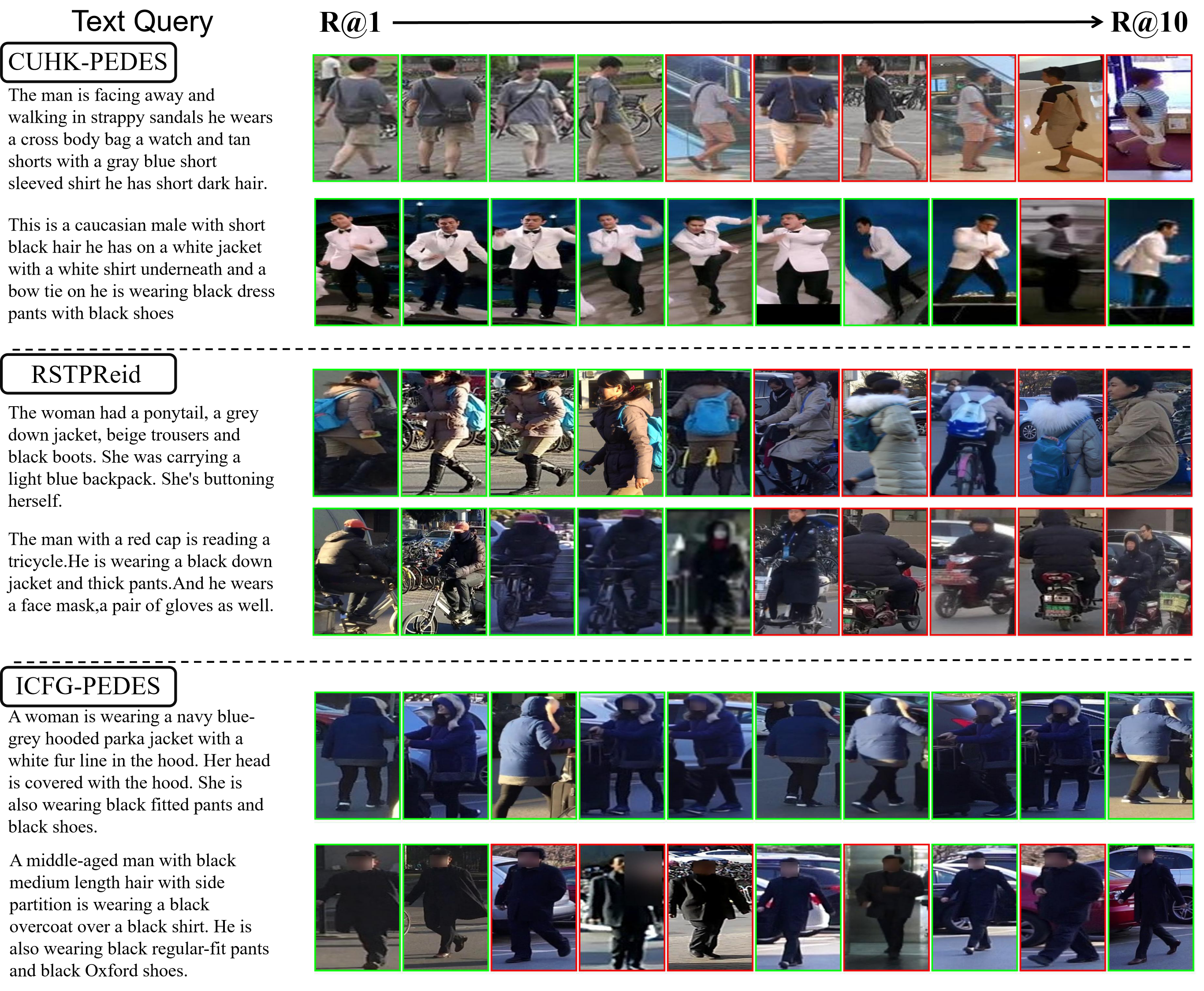}
  \vspace{-2em}
  \caption{ Visualization of the top 10 person search results on CUHK-PEDES, RSTPReid, and ICFG-PEDES. We present the results of two sets of text queries for each of the three datasets, arranged from top to bottom according to our method, in descending order based on match probability. Images in {\color{ForestGreen}{\textbf{green}}} boxes indicate correct matches, while images in {\color{Red}{\textbf{red}}} boxes represent incorrect matches.
  }
  \label{fig:cuhk}
  \vspace{-1em}
\end{figure}

\begin{figure}[t]
  \centering
  \vspace{-4em}
  \includegraphics[width=0.7\linewidth]{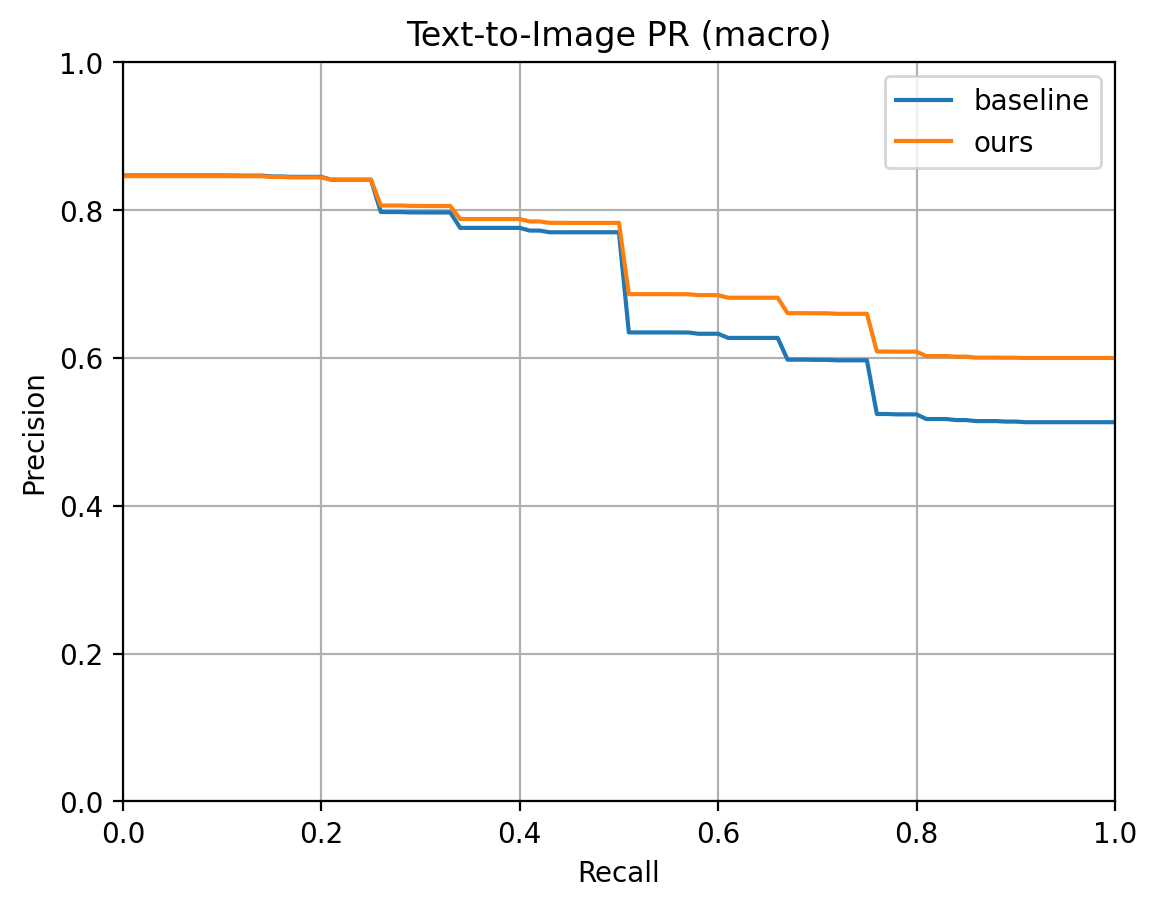}
  \vspace{-1em}
  \caption{Macro-averaged Precision--Recall (PR) curves on CUHK-PEDES for text-to-image retrieval. 
  Our method consistently dominates the baseline across recall levels, indicating improved ranking quality beyond top-K recall. 
  Positives are defined by the same-ID mapping (txt2img) used in mAP evaluation.}
  \label{fig:pr_curve}
  \vspace{-1em}
\end{figure}

\begin{figure}[t]
    \centering
    \vspace{-3.0em}
    \subfloat[Risk--coverage curve by sorting queries with $u_w$ (lower is more reliable).]{
        \includegraphics[width=0.48\linewidth]{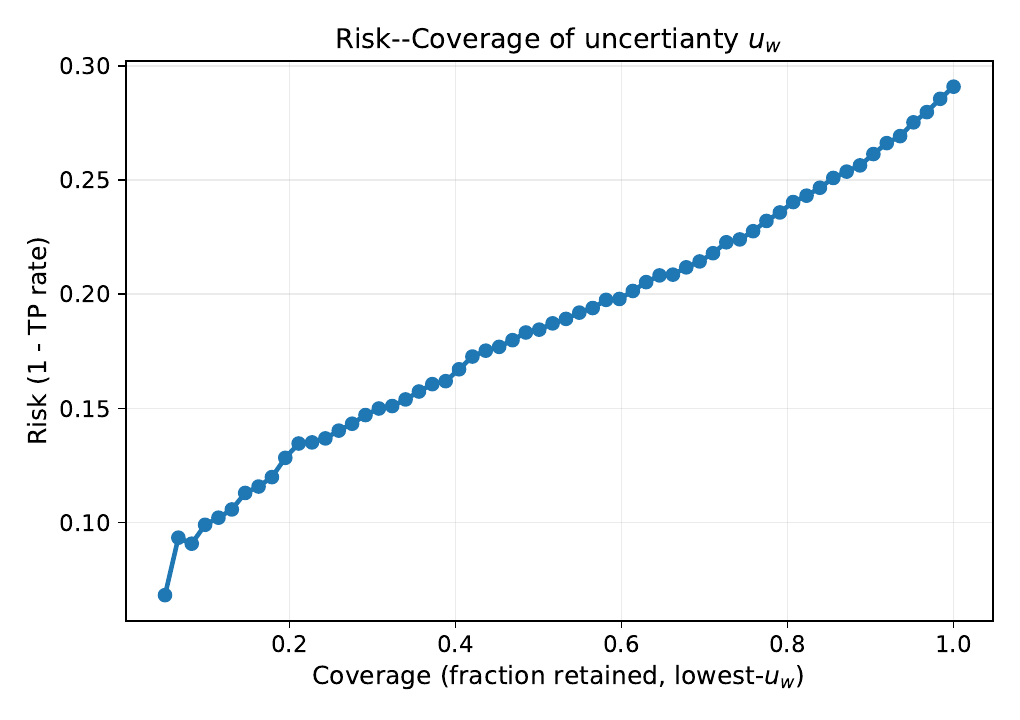}
        \label{fig:uncertainty_rc}
    }\hfill
    \subfloat[Distribution of $u_W$ for correct vs.\ incorrect top-1 retrievals.]{
        \includegraphics[width=0.48\linewidth]{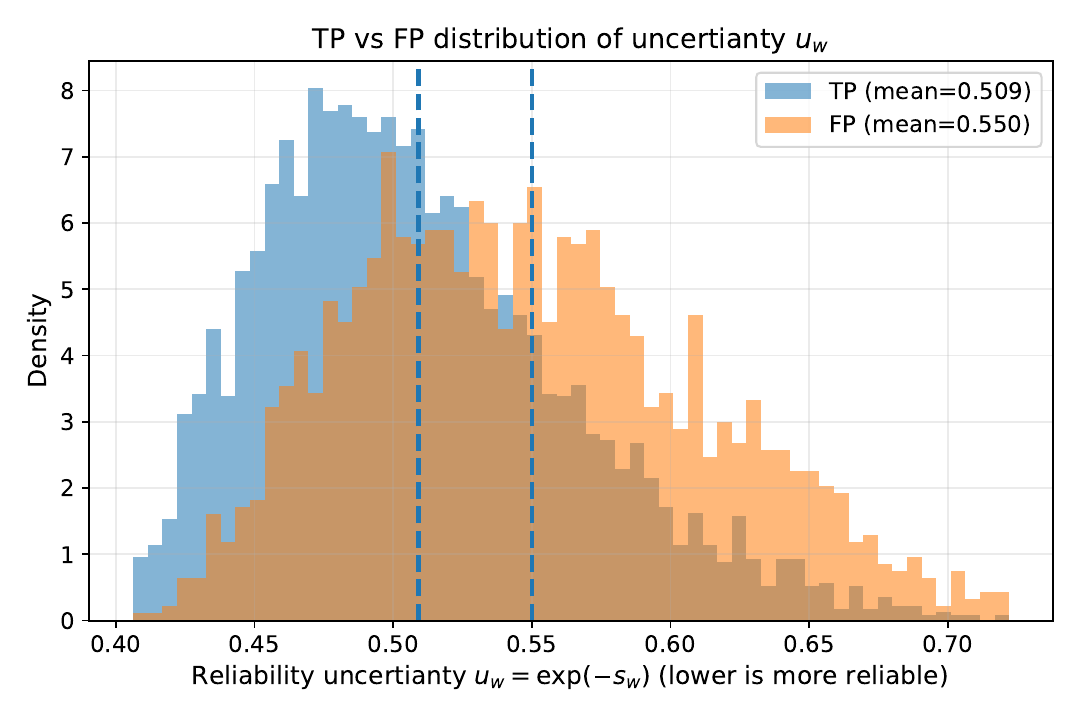}
        \label{fig:uncertainty_dist}
    }
\vspace{-1em}
    \caption{\textbf{Reliability analysis of the consistency-based uncertainty.}
    We emphasize that $u_{w}=\exp(-s_w)$ is a \emph{reliability/ambiguity proxy} derived from weak-pair consistency, rather than a calibrated aleatoric/epistemic uncertainty.
    On CUHK-PEDES test set ($N{=}6156$), incorrect top-1 matches show higher $u$ than correct ones (TP mean $0.509$ vs.\ FP mean $0.550$), and sorting by $u$ yields a risk--coverage behavior, indicating that $u$ is monotonic with retrieval risk.}
    \label{fig:uncertainty_reliability}
\vspace{-1.0em}
\end{figure}

\begin{figure}[t]
\vspace{-6.0em}
    \centering
    \subfloat[\textbf{Joint t-SNE (Before).}]{%
        \includegraphics[width=0.485\linewidth]{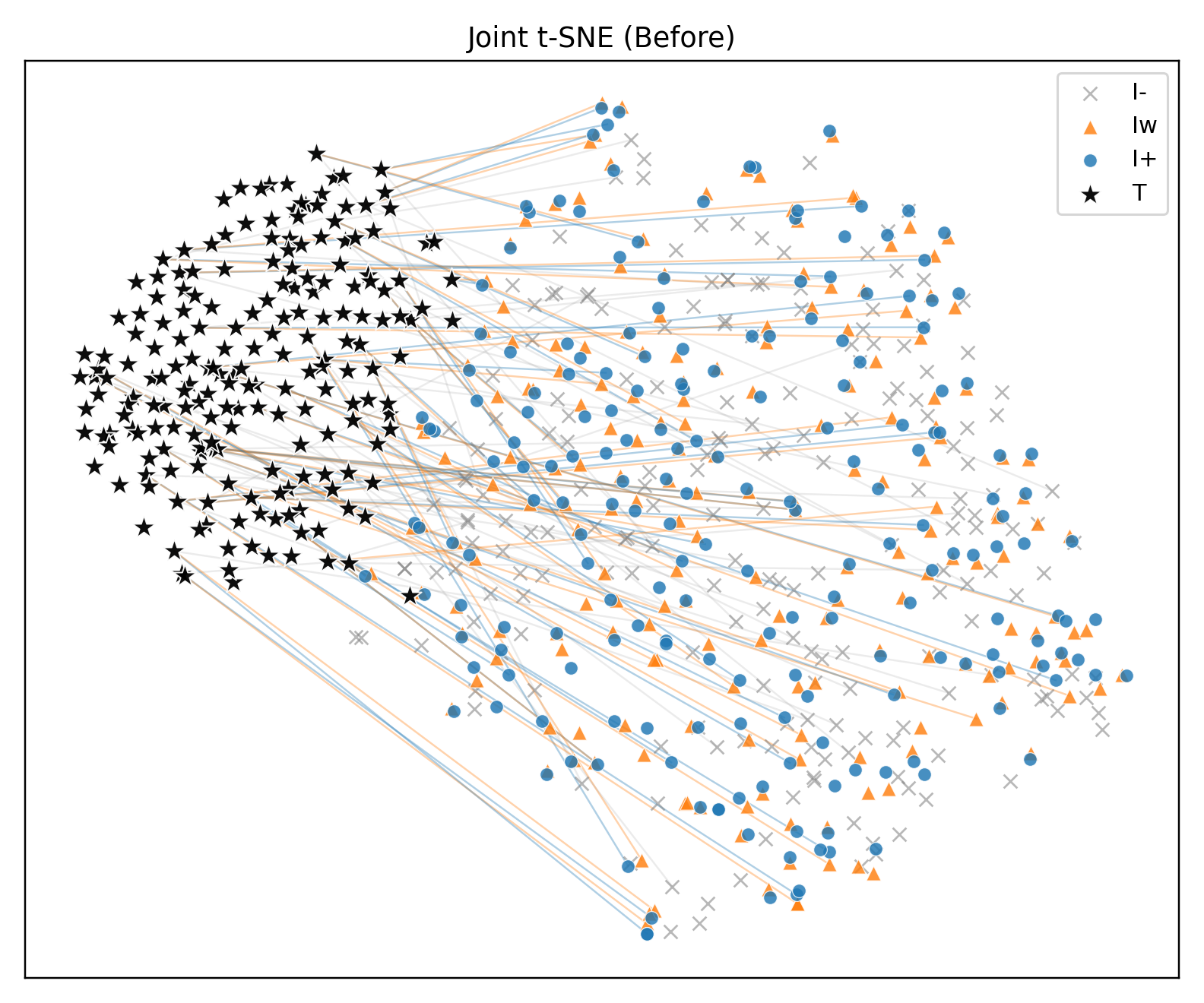}%
        \label{fig:tsne_before}
    }\hfill
    \subfloat[\textbf{Joint t-SNE (After).}]{%
        \includegraphics[width=0.485\linewidth]{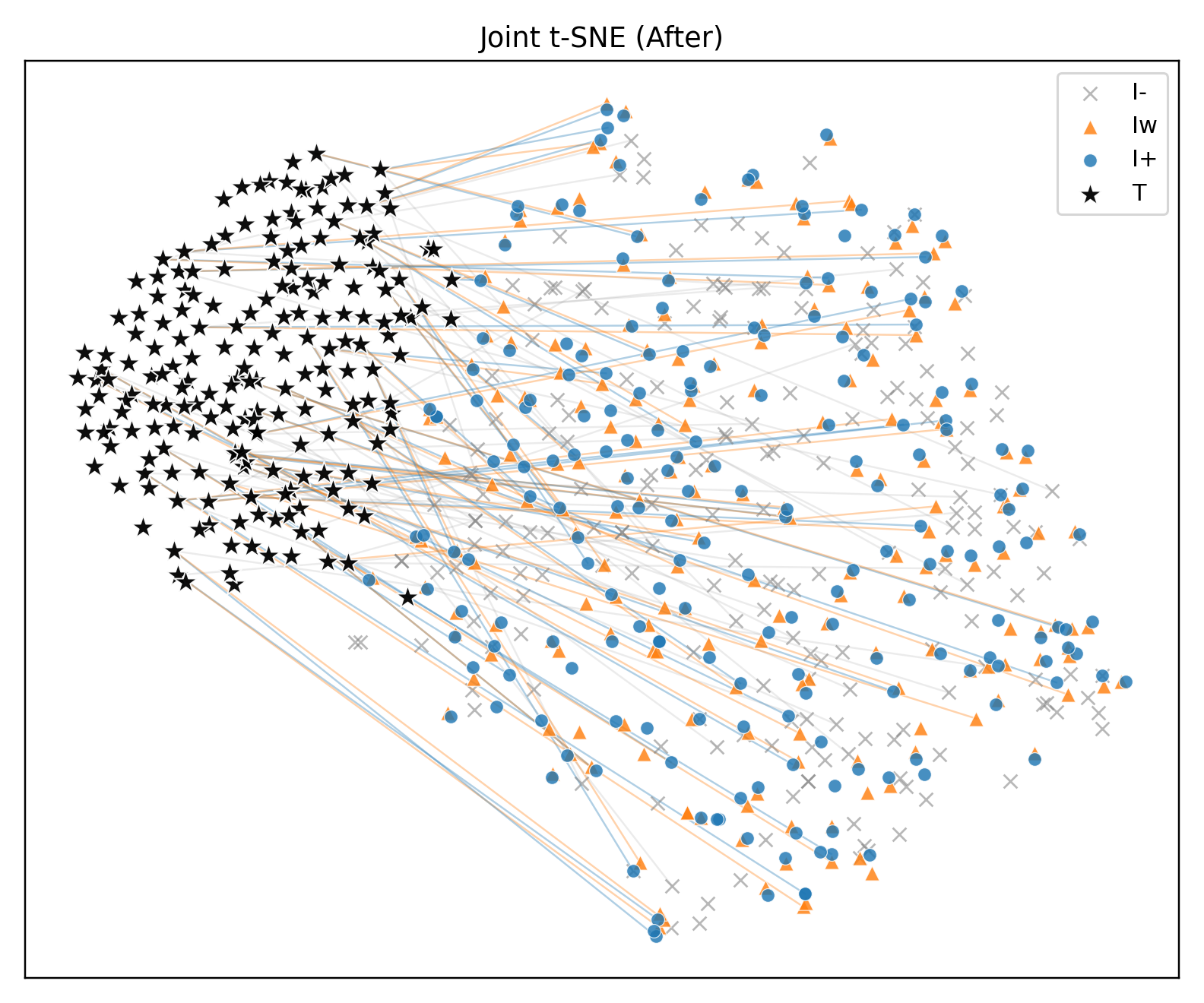}%
        \label{fig:tsne_after}
    }\\
\vspace{-1em}
    \subfloat[\textbf{Weak-positive margin.} $s(T,I_w)-s(T,I^{-})$.]{%
        \includegraphics[width=0.485\linewidth]{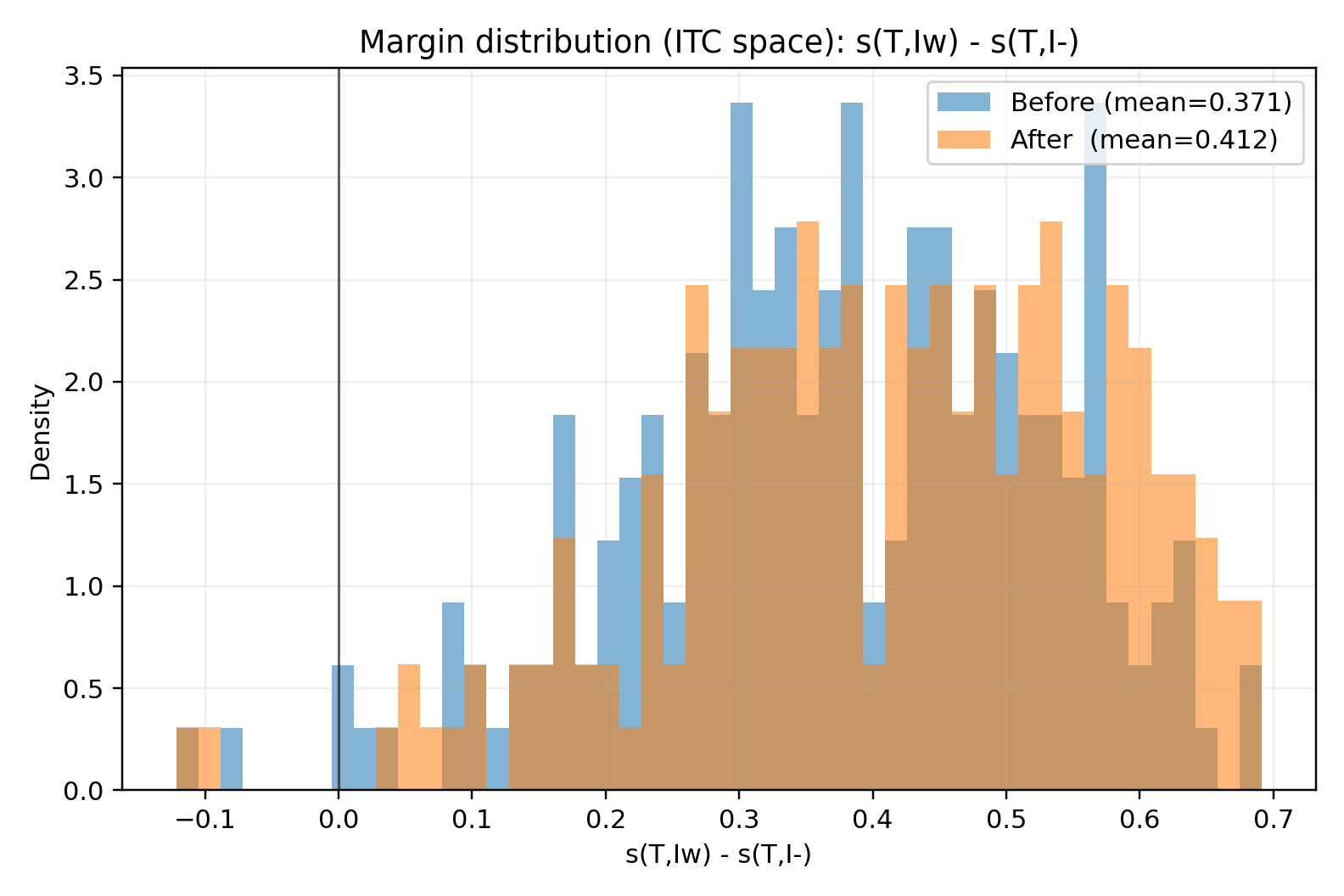}%
        \label{fig:margin_weak}
    }\hfill
    \subfloat[\textbf{Positive margin.} $s(T,I^{+})-s(T,I^{-})$.]{%
        \includegraphics[width=0.485\linewidth]{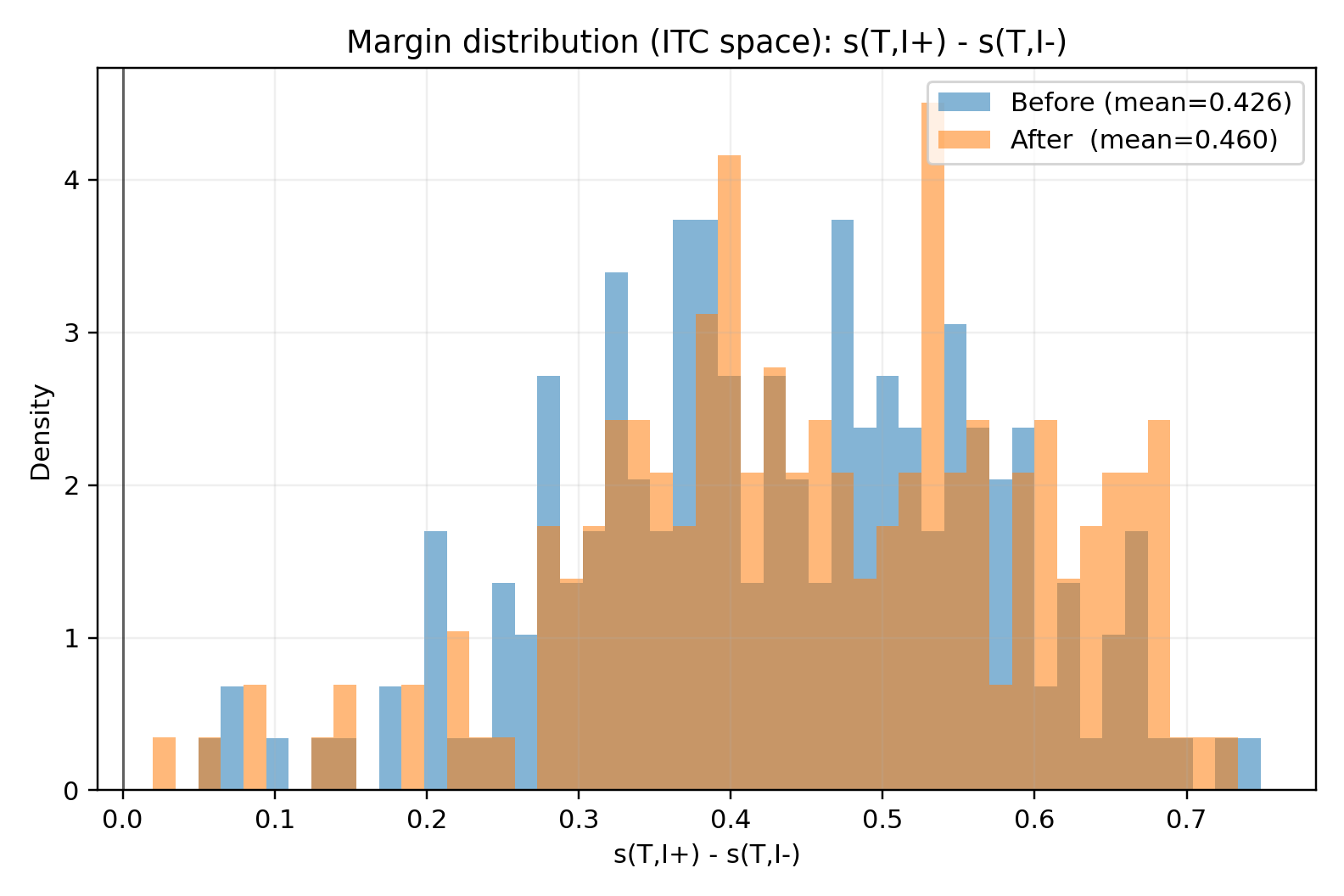}%
        \label{fig:margin_pos}
    }
\vspace{-1.2em}
    \caption{\textbf{Embedding geometry and margin analysis before/after applying our uncertainty-aware learning.}
    (a,b) Joint t-SNE visualization in a shared setting, where $\{T, I^{+}, I_w, I^{-}\}$ denote the text query, its matched image, its weak-view counterpart, and a negative image, respectively. Colored connectors indicate the associations from each query to its positives/weak-positives.
    (c,d) Distributions of ITC margins in the same embedding space. The vertical line at $0$ marks the decision boundary where a negative becomes as similar as (or more similar than) the positive/weak-positive. After training, both margin distributions shift right (larger mean margins), indicating more reliable separation against negatives.}
    \label{fig:tsne_margin}
    \vspace{-1em}
\end{figure}


\textbf{Impact of GITM Loss Weight $\beta$.}
We further study the sensitivity of our framework to the weighting coefficient $\beta$ for the GITM loss by sweeping a wider range from $0.01$ to $0.5$ on CUHK-PEDES while fixing $\alpha=0.1$. As shown in Table~\ref{tab:beta}, the performance is relatively robust to $\beta$ within this range, yet different metrics favor slightly different choices. In particular, $\beta=0.1$ attains the best R@1 ($77.88\%$) with competitive mAP ($72.44\%$), whereas the highest mAP is achieved at $\beta=0.2$ ($72.47\%$). For smaller weights, $\beta=0.05$ yields the best R@5 ($91.16\%$) and a strong mAP ($71.68\%$), and $\beta=0.01$ gives the best R@10 ($94.74\%$) but a lower mAP ($70.23\%$). When $\beta$ becomes larger (e.g., $\ge 0.3$), Recall@K consistently degrades (from $77.88\%$ R@1 at $\beta=0.1$ to $76.07\%$ at $\beta=0.4$), while mAP remains nearly saturated around $72.0\%$--$72.1\%$. Overall, these results indicate that GITM mainly acts as an auxiliary regularizer: overly increasing its weight does not bring additional benefits and can slightly compromise retrieval recall. Therefore, we adopt $\beta=0.1$ as the default setting in our experiments.

\textbf{Comparing the impact of different input sizes in image backbone.}
We consider the size of the input image and window size to perceive the model receptive field and learning details. 
All experimental results are obtained after applying all uncertainty methods. For specific results, refer to Table~\ref{tab:ab3}. It can be seen from the table that using 384 $\times$ 384 as H $\times$ W for image processing for the three existing datasets will achieve the best results in the case of Recall@1, 5, 10, and mAP. 
Additionally, we observe that reducing both the height and width of the images to 256 $\times$ 256 leads to a decline in all performance metrics. Similarly, keeping the height constant while reducing the width to 192 results in a performance drop, particularly in mAP. On the other hand, increasing the height to 576 while keeping the width at 192 causes a decrease in R@1 and R@5, but an improvement in R@10 and mAP.

\begin{table}[t]
\centering
\footnotesize
\setlength{\tabcolsep}{6pt}
\renewcommand{\arraystretch}{1.1}
\caption{Comparison of different uncertainty mappings in Eq.~(3) on CUHK-PEDES. All settings are kept identical, and we only replace the mapping from the consistency score $s_w$ to the uncertainty score $u_w$.}
\label{tab:mapping_ablation}
\begin{tabular}{lcccc}
\toprule
Mapping of $u_w$ & R@1 & R@5 & R@10 & mAP \\
\midrule
$1.5 - s_w$ & 77.00 & 90.48 & 94.14 & 71.88 \\
$(1.5 - s_w)^2$ & 77.73 & 90.60 & 93.96 & 72.35 \\
$\exp(-s_w)$ & \textbf{77.88} & \textbf{91.05} & \textbf{94.57} & \textbf{72.44} \\
\bottomrule
\end{tabular}
\end{table}

\textbf{Comparison of uncertainty mappings.}
The uncertainty score in Eq.~(3) is instantiated as $u_w=\exp(-s_w)$ in our default setting. Since $s_w$ is computed from cosine similarities on L2-normalized features, it is bounded in $[-1,1]$, and thus the exponential mapping keeps $u_w$ strictly positive and bounded, which is desirable because $u_w$ appears in the denominator of Eq.~(4). To further examine whether this choice is empirically reasonable, we compare it with two simple positive monotonic alternatives, i.e., a linear mapping $u_w=1.5-s_w$ and a power-based mapping $u_w=(1.5-s_w)^2$. As shown in Table~\ref{tab:mapping_ablation}, the exponential mapping achieves the best overall performance on CUHK-PEDES, outperforming the linear variant by +0.88 R@1 and +0.56 mAP, and also slightly surpassing the power-based variant by +0.15 R@1 and +0.09 mAP. These results suggest that the exponential form provides a more suitable non-linear reweighting for low-consistency weak pairs in our current objective. We emphasize that we do not claim $\exp(-s_w)$ to be universally optimal; rather, it is a stable, simple, and empirically effective choice for the present uncertainty-aware ITC formulation.

\begin{table}[t]
\centering
\footnotesize
\setlength{\tabcolsep}{6pt}
\renewcommand{\arraystretch}{1.08}
\caption{Effect of introducing pair-level uncertainty into GITM on CUHK-PEDES. ``Default'' denotes our original design, where $u_w$ is only used in the ITC branch. ``Uncertainty-weighted GITM'' additionally applies pair-level uncertainty weighting to the weak-positive GITM branches.}
\label{tab:gitm_unc_ablation}
\begin{tabular}{lcccc}
\toprule
\textbf{GITM supervision} & \textbf{R@1} & \textbf{R@5} & \textbf{R@10} & \textbf{mAP} \\
\midrule
Default (ours) & \textbf{77.88} & \textbf{91.05} & \textbf{94.57} & \textbf{72.44} \\
Uncertainty-weighted GITM & 77.27 & 90.48 & 94.19 & 72.40 \\
\bottomrule
\end{tabular}
\end{table}

\textbf{Does GITM also benefit from uncertainty weighting?}
To further examine whether $u_w$ should also be introduced into GITM, we implement a variant that applies pair-level uncertainty weighting to the weak-positive GITM branches. As shown in Table~\ref{tab:gitm_unc_ablation}, this modification does not bring further improvement over the default design: Recall@1/5/10 drop from 77.88/91.05/94.57 to 77.27/90.48/94.19, while mAP changes only marginally from 72.44 to 72.40. This result suggests that uncertainty-aware weighting is more suitable for the ITC branch, where it continuously regulates pair-level ambiguity in the contrastive space, whereas GITM is more effective as an auxiliary binary matching branch whose main gain comes from enriching the group structure and increasing the number and difficulty of hard negatives.

\begin{table}[t]
\centering
\footnotesize
\setlength{\tabcolsep}{6pt}
\renewcommand{\arraystretch}{1.08}
\caption{Effect of removing MALS pre-training on CUHK-PEDES. Here ``w/o MALS pre-training'' means that we keep the same APTM architecture and standard backbone initialization, but directly train on the downstream dataset without loading the MALS pre-trained checkpoint.}
\label{tab:wo_mals}
\begin{tabular}{lcccc}
\toprule
\textbf{Method} & \textbf{R@1} & \textbf{R@5} & \textbf{R@10} & \textbf{mAP} \\
\midrule
Baseline w/o MALS pre-training & 68.81 & 86.60 & 91.42 & 62.36 \\
Ours w/o MALS pre-training & \textbf{70.37} & \textbf{87.12} & \textbf{91.96} & \textbf{66.20} \\
\bottomrule
\end{tabular}
\end{table}

\textbf{Does our method depend on MALS pre-training?}
Since our approach is built on the standard pretrain--finetune protocol of APTM, one may ask whether the uncertainty estimation heavily depends on the large-scale MALS pre-trained feature space. To examine this, we conduct an additional experiment without MALS pre-training. Specifically, we keep the same APTM architecture and the standard backbone initialization, but directly train on CUHK-PEDES without loading the MALS pre-trained checkpoint. As shown in Table~\ref{tab:wo_mals}, the baseline achieves 68.81 R@1 and 62.36 mAP, while our full method further improves the performance to 70.37 R@1 and 66.20 mAP. This corresponds to gains of +1.56 R@1 and +3.84 mAP. These results suggest that, although MALS pre-training provides a stronger starting point, the proposed uncertainty-aware learning and GITM do not rely on it to remain effective. The uncertainty signal is still computed online from the current feature space and can provide useful supervision beyond the original MALS-pretrained setting.

\subsection{Qualitative Results}~\label{visual}
As shown in Fig~\ref{fig:cuhk}, we provide qualitative results of the top 10 search results on three datasets: CUHK-PEDES, RSTPReid, and ICFG-PEDES. Our model uses uncertainty learning to improve precision compared to the baseline. In addition, compared with the conventional baseline approach of contrast learning, we observe that the proposed uncertainty adjustment has better recognition for small-scale targets such as backpacks and tote bags. This is because many descriptions of the same ID have different perspectives, and some perspectives obscure objects such as hand-held objects or backpacks, which makes model learning and text description have limitations. Our uncertainty-aware method corrects these biases and gets reasonable search results.

\textbf{Ranking-level PR analysis.}
\label{sec:pr_analysis}
We further provide a ranking-level diagnostic by plotting the macro-averaged Precision--Recall (PR) curve for text-to-image retrieval on CUHK-PEDES (Fig.~\ref{fig:pr_curve}).
For each text query, we rank all gallery images using the final retrieval score (the same score used for mAP evaluation) and define positives as all images sharing the same identity (consistent with our evaluation protocol via \texttt{txt2img}).
As shown in Fig.~\ref{fig:pr_curve}, our method dominates the baseline across most recall levels, indicating improved ranking quality beyond top-$K$ recall.
Quantitatively, Precision@Recall improves from $0.770/0.597/0.514$ to $0.783/0.660/0.600$ at recall $=0.5/0.7/0.9$, and PR-AUC increases from $0.692$ to $0.730$, with more pronounced gains in the mid-to-high recall regime.

\textbf{Consistency-based uncertainty reliability analysis.}
We clarify that our uncertainty is a retrieval-oriented score derived from cross-view (weak-pair) consistency, rather than a probabilistically calibrated aleatoric/epistemic estimate.
To validate its reliability meaning, we perform a diagnostic on CUHK-PEDES test set ($N{=}6156$).
We observe that incorrect top-1 retrievals exhibit higher uncertainty than correct ones (TP mean $0.509$ vs.\ FP mean $0.550$), and the resulting risk--coverage curve shows that retaining the lowest-uncertainty fraction substantially reduces error (Fig.~\ref{fig:uncertainty_reliability}).
These results support that the proposed uncertainty is monotonic with retrieval risk and thus suitable for reliability-aware optimization.

\textbf{Embedding geometry analysis.}
To further understand how the proposed uncertainty-aware learning reshapes the joint embedding space, we visualize the representations of sampled tuples $\{T, I^{+}, I_w, I^{-}\}$, where $I_w$ denotes a weak-view counterpart under the same identity.
As shown in Fig.~\ref{fig:tsne_margin}(a--b), compared with the baseline, our method yields a visibly more coherent text--image structure: the associations from $T$ to $I^{+}$ and $I_w$ exhibit fewer cross-cluster jumps, suggesting improved cross-view consistency and reduced ambiguity under weak-view perturbations.
We complement this qualitative evidence with a margin-based analysis in the ITC space. Fig.~\ref{fig:tsne_margin}(c--d) reports the distributions of the margins $s(T,I_w)-s(T,I^{-})$ and $s(T,I^{+})-s(T,I^{-})$, where the $0$-line indicates cases where negatives become competitive.
After training, both margin distributions shift toward larger values (higher mean margins), indicating that our learning strategy increases the safety margin against negatives not only for the strongest positive pairs $(T,I^{+})$ but also for weak-view pairs $(T,I_w)$, which is consistent with the goal of improving reliability under cross-view variations.

\section{Conclusion}~\label{conclusion}
In this work, we propose a simple and effective method to improve the text-based person search by harnessing the weak positive pairs. We apply uncertainty in the cross-modality comparison and incorporate it into the adjustment of loss learning to correct the training. We further introduce group-wise image text matching to enhance metric learning. Therefore, our method presents an attempt to motivate the model to fully leverage the information of image text pairs without introducing additional parameters and modules. 
We achieve competitive performance on three benchmarks, and a large number of experiments indicate the effectiveness of the proposed approach in text-based person search, especially in terms of precision. In the future, we will continue to investigate the utilization of uncertainty and its application to other related tasks.











\bibliography{mybibfile}

@String(CVPR= {IEEE Conf. Comput. Vis. Pattern Recog.})

@String(ICCV= {Int. Conf. Comput. Vis.})

@String(ECCV= {Eur. Conf. Comput. Vis.})

@String(ACCV  = {ACCV})

@String(IJCAI = {IJCAI})

@String(AAAI = {AAAI})

@inproceedings{APTM,
  title={Towards Unified Text-based Person Retrieval: A Large-scale Multi-Attribute and Language Search Benchmark},
  author={Yang, Shuyu and Zhou, Yinan and Wang, Yaxiong and Wu, Yujiao and Zhu, Li and Zheng, Zhedong},
  booktitle = {Proceedings of the 2023 {ACM} on Multimedia Conference},
  year={2023}
}

@inproceedings{SwinTransformer,  
 title={Swin Transformer: Hierarchical Vision Transformer using Shifted Windows}, 
 DOI={10.1109/iccv48922.2021.00986}, 
 booktitle={ICCV}, 
 author={Liu, Ze and Lin, Yutong and Cao, Yue and Hu, Han and Wei, Yixuan and Zhang, Zheng and Lin, Stephen and Guo, Baining}, 
 year={2021}
 }

@inproceedings{Li_Xiao_Li_Zhou_Yue_Wang_2017,  
 title={Person Search with Natural Language Description}, 
 DOI={10.1109/cvpr.2017.551}, 
 booktitle={CVPR}, 
 author={Li, Shuang and Xiao, Tong and Li, Hongsheng and Zhou, Bolei and Yue, Dayu and Wang, Xiaogang}, 
 year={2017}, 
 month={Jul}, 
 language={en-US} 
 }

@inproceedings{Shao_Zhang_Fang_Lin_Wang_Ding_2022,
author = {Shao, Zhiyin and Zhang, Xinyu and Fang, Meng and Lin, Zhifeng and Wang, Jian and Ding, Changxing},
title = {Learning Granularity-Unified Representations for Text-to-Image Person Re-identification},
year = {2022},
isbn = {9781450392037},
publisher = {Association for Computing Machinery},
address = {New York, NY, USA},
doi = {10.1145/3503161.3548028},
booktitle = {ACM MM},
pages = {5566–5574},
numpages = {9}
}

@inproceedings{Wang_Zhu_Xue_Wan_Liu_Wang_Li_2022,  
 title={Look Before You Leap: Improving Text-based Person Retrieval by Learning A Consistent Cross-modal Common Manifold}, 
 DOI={10.1145/3503161.3548166}, 
 booktitle={ACM MM}, 
 author={Wang, Zijie and Zhu, Aichun and Xue, Jingyi and Wan, Xili and Liu, Chao and Wang, Tian and Li, Yifeng}, 
 year={2022}
 }

@article{Chen_Zhang_Lu_Wang_Zheng_2022,  
 title={TIPCB: A simple but effective part-based convolutional baseline for text-based person search}, 
 DOI={10.1016/j.neucom.2022.04.081}, 
 journal={Neurocomputing}, 
 author={Chen, Yuhao and Zhang, Guoqing and Lu, Yujiang and Wang, Zhenxing and Zheng, Yuhui}, 
 year={2022}
 }

@article{Ding_Ding_Shao_Tao_2021,  
 title={Semantically Self-Aligned Network for Text-to-Image Part-aware Person Re-identification.}, 
 journal={arXiv}, 
 author={Ding, Zefeng and Ding, Changxing and Shao, Zhiyin and Tao, Dacheng}, 
 year={2021}
 }

@inproceedings{CAIBC,  
 title={CAIBC: Capturing All-round Information Beyond Color for Text-based Person Retrieval}, 
 DOI={10.1145/3503161.3548057}, 
 booktitle={ACM MM}, 
 author={Wang, Zijie and Zhu, Aichun and Xue, Jingyi and Wan, Xili and Liu, Chao and Wang, Tian and Li, Yifeng}, 
 year={2022}
 }

@article{Zheng_Zheng_Garrett_Yang_Xu_Shen_2020,  
 title={Dual-Path Convolutional Image-Text Embedding with Instance Loss}, 
 DOI={10.1145/3383184}, 
 journal={ACM Transactions on Multimedia Computing, Communications, and Applications}, 
 author={Zheng, Zhedong and Zheng, Liang and Garrett, Michael and Yang, Yi and Xu, Mingliang and Shen, Yi-Dong}, 
 year={2020}, 
 pages={1–23}
 }

@article{niu2019improving,
  title={Improving description-based person re-identification by multi-granularity image-text alignments},
  author={Niu, Kai and Huang, Yan and Ouyang, Wanli and Wang, Liang},
  journal={IEEE Transactions on Image Processing},
  volume={29},
  pages={5542--5556},
  year={2020},
  publisher={IEEE}
}

@ARTICLE{10345496,
  author={He, Shuting and Luo, Hao and Jiang, Wei and Jiang, Xudong and Ding, Henghui},
  journal={IEEE Transactions on Image Processing}, 
  title={{VGSG}: Vision-Guided Semantic-Group Network for Text-Based Person Search}, 
  year={2024},
  volume={33},
  number={},
  pages={163-176},
  doi={10.1109/TIP.2023.3337653}}

@inproceedings{Bai_2023, series={IJCAI-2023},
   title={RaSa: Relation and Sensitivity Aware Representation Learning for Text-based Person Search},
   DOI={10.24963/ijcai.2023/62},
   booktitle={IJCAI},
   author={Bai, Yang and Cao, Min and Gao, Daming and Cao, Ziqiang and Chen, Chen and Fan, Zhenfeng and Nie, Liqiang and Zhang, Min},
   year={2023}
}

@article{Xiao_Li_Wang_Li_Wang_2016,  
 title={End-to-End Deep Learning for Person Search.}, 
 journal={arXiv}, 
 author={Xiao, Tong and Li, Shuang and Wang, Bochao and Li, Lin and Wang, Xiaogang}, 
 year={2016}, 
 month={Apr}, 
 language={en-US} 
 }

@misc{devlin2019bert,
      title={BERT: Pre-training of Deep Bidirectional Transformers for Language Understanding}, 
      author={Jacob Devlin and Ming-Wei Chang and Kenton Lee and Kristina Toutanova},
      year={2019},
      eprint={1810.04805},
      archivePrefix={arXiv},
      primaryClass={cs.CL}
}

@inproceedings{Wei_Zhang_Gao_Tian_2018,  
 title={Person Transfer GAN to Bridge Domain Gap for Person Re-Identification}, 
 DOI={10.1109/cvpr.2018.00016}, 
 booktitle={CVPR}, 
 author={Wei, Longhui and Zhang, Shiliang and Gao, Wen and Tian, Qi}, 
 year={2018}
 }

@inproceedings{Li_Zhao_Xiao_Wang_2014,  
 title={DeepReID: Deep Filter Pairing Neural Network for Person Re-identification}, 
 DOI={10.1109/cvpr.2014.27}, 
 booktitle={2014 IEEE Conference on Computer Vision and Pattern Recognition}, 
 author={Li, Wei and Zhao, Rui and Xiao, Tong and Wang, Xiaogang}, 
 year={2014}, 
 language={en-US} 
 }

@article{Zheng_Shen_Tian_Wang_Bu_Tian_2015,  
 title={Person Re-identification Meets Image Search}, 
 journal={arXiv}, 
 author={Zheng, Liang and Shen, Liyue and Tian, Lei and Wang, Shengjin and Bu, Jiahao and Tian, Qi}, 
 year={2015},
 language={en-US} 
 }

@inproceedings{Gray_Brennan_Tao_2007,
  title={Evaluating appearance models for recognition, reacquisition, and tracking},
  author={Gray, Douglas and Brennan, Shane and Tao, Hai},
  booktitle={PETS},
  year={2007}
}

@inproceedings{Li_Zhao_Wang_2013,
  title={Human reidentification with transferred metric learning},
  author={Li, Wei and Zhao, Rui and Wang, Xiaogang},
  booktitle={ACCV},
  pages={31--44},
  year={2013},
  organization={Springer}
}

@article{Loshchilov_Hutter_2017,  
 title={Decoupled Weight Decay Regularization}, 
 journal={Learning,Learning}, 
 author={Loshchilov, Ilya and Hutter, Frank}, 
 year={2017}, 
 month={Nov}, 
 language={en-US} 
 }

@inproceedings{Cubuk_Zoph_Shlens_Le_2020,  
 title={Randaugment: Practical automated data augmentation with a reduced search space}, 
 DOI={10.1109/cvprw50498.2020.00359}, 
 booktitle={CVPR Workshop}, 
 author={Cubuk, Ekin D. and Zoph, Barret and Shlens, Jonathon and Le, Quoc V.}, 
 year={2020}, 
 month={Jun}, 
 language={en-US} 
 }

@article{Zhong_Zheng_Kang_Li_Yang_2020,  
 title={Random Erasing Data Augmentation}, 
 DOI={10.1609/aaai.v34i07.7000}, 
 journal={AAAI}, 
 author={Zhong, Zhun and Zheng, Liang and Kang, Guoliang and Li, Shaozi and Yang, Yi}, 
 year={2020}, 
 month={Jun}, 
 pages={13001–13008}, 
 language={en-US} 
 }

@inproceedings{Wei_Zou_2019,  
 title={EDA: Easy Data Augmentation Techniques for Boosting Performance on Text Classification Tasks}, 
 DOI={10.18653/v1/d19-1670}, 
 booktitle={EMNLP-IJCNLP}, 
 author={Wei, Jason and Zou, Kai}, 
 year={2019}
 }

@inproceedings{lee2018stacked,
  title={Stacked cross attention for image-text matching},
  author={Lee, Kuang-Huei and Chen, Xi and Hua, Gang and Hu, Houdong and He, Xiaodong},
  booktitle={ECCV},
  year={2018}
}

@inproceedings{jiang2023crossmodal,
  title={Cross-Modal Implicit Relation Reasoning and Aligning for Text-to-Image Person Retrieval},
  author={Jiang, Ding and Ye, Mang},
  booktitle={CVPR},
  year={2023},
}

@ARTICLE{yan2022clipdriven,
  author={Yan, Shuanglin and Dong, Neng and Zhang, Liyan and Tang, Jinhui},
  journal={IEEE Transactions on Image Processing}, 
  title={{CLIP}-Driven Fine-Grained Text-Image Person Re-Identification}, 
  year={2023},
  volume={32},
  number={},
  pages={6032-6046},
  doi={10.1109/TIP.2023.3327924}}

@inproceedings{zhu2021dssl,
author = {Zhu, Aichun and Wang, Zijie and Li, Yifeng and Wan, Xili and Jin, Jing and Wang, Tian and Hu, Fangqiang and Hua, Gang},
title = {DSSL: Deep Surroundings-person Separation Learning for Text-based Person Retrieval},
year = {2021},
isbn = {9781450386517},
publisher = {Association for Computing Machinery},
address = {New York, NY, USA},
doi = {10.1145/3474085.3475369},
booktitle = {ACM Multimedia},
pages = {209–217},
numpages = {9},
series = {MM '21}
}

@inproceedings{cao2023empirical,
author = {Cao, Min and Bai, Yang and Zeng, Ziyin and Ye, Mang and Zhang, Min},
title = {An empirical study of {{CLIP}} for text-based person search},
year = {2024},
isbn = {978-1-57735-887-9},
publisher = {AAAI},
doi = {10.1609/aaai.v38i1.27801},
articleno = {53},
numpages = {9},
}

@inproceedings{oursMM,
  title={From data deluge to data curation: A filtering-wora paradigm for efficient text-based person search},
  author={Sun, Jintao and Fei, Hao and Ding, Gangyi and Zheng, Zhedong},
  booktitle={ACM WWW},
  year={2025}
}

@article{Kendall_Gal_2017,
  title={What uncertainties do we need in bayesian deep learning for computer vision?},
  author={Kendall, Alex and Gal, Yarin},
  journal={Advances in neural information processing systems},
  volume={30},
  year={2017}
}

@inproceedings{Warburg_Jorgensen_Civera_Hauberg_2021,  
 title={Bayesian Triplet Loss: Uncertainty Quantification in Image Retrieval}, 
 DOI={10.1109/iccv48922.2021.01194}, 
 booktitle={ICCV}, 
 author={Warburg, Frederik and Jorgensen, Martin and Civera, Javier and Hauberg, Soren}, 
 year={2021}, 
 month={Oct}, 
 language={en-US} 
 }

@article{Postels_Segu_Sun_Gool_Yu_Tombari_2021,  
 title={On the Practicality of Deterministic Epistemic Uncertainty.}, 
 journal={arXiv}, 
 author={Postels, Janis and Segu, Mattia and Sun, Tao and Gool, LucVan and Yu, Fisher and Tombari, Federico}, 
 year={2021}, 
 month={Jul}, 
 language={en-US} 
 }

@article{10.1007/s11263-020-01395-y,
author = {Zheng, Zhedong and Yang, Yi},
title = {Rectifying Pseudo Label Learning via Uncertainty Estimation for Domain Adaptive Semantic Segmentation},
year = {2021},
issue_date = {Apr 2021},
publisher = {Kluwer Academic Publishers},
address = {USA},
volume = {129},
number = {4},
issn = {0920-5691},
doi = {10.1007/s11263-020-01395-y},
journal = {Int. J. Comput. Vision},
month = {apr},
pages = {1106–1120},
numpages = {15}
}

@misc{chen2024composed,
      title={Composed Image Retrieval with Text Feedback via Multi-grained Uncertainty Regularization}, 
      author={Yiyang Chen and Zhedong Zheng and Wei Ji and Leigang Qu and Tat-Seng Chua},
      year={2024},
      eprint={2211.07394},
      archivePrefix={arXiv},
      primaryClass={cs.CV}
}

@inproceedings{Chang_Lan_Cheng_Wei_2020,  
 title={Data Uncertainty Learning in Face Recognition}, 
 DOI={10.1109/cvpr42600.2020.00575}, 
 booktitle={CVPR}, 
 author={Chang, Jie and Lan, Zhonghao and Cheng, Changmao and Wei, Yichen}, 
 year={2020}, 
 month={Jun}, 
 language={en-US} 
 }

@misc{dou2022reliabilityaware,
      title={Reliability-Aware Prediction via Uncertainty Learning for Person Image Retrieval}, 
      author={Zhaopeng Dou and Zhongdao Wang and Weihua Chen and Yali Li and Shengjin Wang},
      year={2022},
      eprint={2210.13440},
      archivePrefix={arXiv},
      primaryClass={cs.CV}
}

@misc{oh2019modeling,
      title={Modeling Uncertainty with Hedged Instance Embedding}, 
      author={Seong Joon Oh and Kevin Murphy and Jiyan Pan and Joseph Roth and Florian Schroff and Andrew Gallagher},
      year={2019},
      eprint={1810.00319},
      archivePrefix={arXiv},
      primaryClass={cs.LG}
}

@INPROCEEDINGS{9710535,
  author={Warburg, Frederik and Jørgensen, Martin and Civera, Javier and Hauberg, Søren},
  booktitle={ICCV}, 
  title={Bayesian Triplet Loss: Uncertainty Quantification in Image Retrieval}, 
  year={2021},
  volume={},
  number={},
  pages={12138-12148},
  doi={10.1109/ICCV48922.2021.01194}}

@article{Marchette_2003,  
 title={Bayesian Networks and Decision Graphs}, 
 volume={45}, 
 DOI={10.1198/tech.2003.s141}, 
 number={2}, 
 journal={Technometrics}, 
 author={Marchette, David J}, 
 year={2003}, 
 month={May}, 
 pages={178–179}, 
 language={en-US} 
 }

@inproceedings{Gal_Ghahramani_2015,
  title={Dropout as a bayesian approximation: Representing model uncertainty in deep learning},
  author={Gal, Yarin and Ghahramani, Zoubin},
  booktitle={international conference on machine learning},
  pages={1050--1059},
  year={2016},
  organization={PMLR}
}

@article{Sun2024AnAC,
author = {Sun, Mengyang and Suo, Wei and Wang, Peng and Niu, Kai and Liu, Le and Lin, Guosheng and Zhang, Yanning and Wu, Qi},
title = {An Adaptive Correlation Filtering Method for Text-Based Person Search},
year = {2024},
issue_date = {Oct 2024},
publisher = {Kluwer Academic Publishers},
address = {USA},
volume = {132},
number = {10},
issn = {0920-5691},
doi = {10.1007/s11263-024-02094-8},
journal = {Int. J. Comput. Vision},
month = may,
pages = {4440–4455},
numpages = {16},
}

@inproceedings{ALBEF,
author = {Li, Junnan and Selvaraju, Ramprasaath R. and Gotmare, Akhilesh D. and Joty, Shafiq and Xiong, Caiming and Hoi, Steven C.H.},
title = {Align before fuse: vision and language representation learning with momentum distillation},
year = {2021},
isbn = {9781713845393},
booktitle = {NeurIPS},
articleno = {742},
numpages = {12}
}

@inproceedings{Shu_Wen_Wu_Chen_Song_Qiao_Ren_Wang_2022,
author = {Shu, Xiujun and Wen, Wei and Wu, Haoqian and Chen, Keyu and Song, Yiran and Qiao, Ruizhi and Ren, Bo and Wang, Xiao},
title = {See Finer, See More: Implicit Modality Alignment for Text-based Person Retrieval},
year = {2023},
isbn = {978-3-031-25071-2},
publisher = {Springer-Verlag},
address = {Berlin, Heidelberg},
doi = {10.1007/978-3-031-25072-9_42},
booktitle = {Computer Vision – ECCV 2022 Workshops: Tel Aviv, Israel, October 23–27, 2022, Proceedings, Part V},
pages = {624–641},
numpages = {18},
location = {Tel Aviv, Israel}
}

@ARTICLE{TransTPS,
  author={Bao, Liping and Wei, Longhui and Zhou, Wengang and Liu, Lin and Xie, Lingxi and Li, Houqiang and Tian, Qi},
  journal={IEEE Transactions on Multimedia}, 
  title={Multi-Granularity Matching Transformer for Text-Based Person Search}, 
  year={2024},
  volume={26},
  number={},
  pages={4281-4293},
  doi={10.1109/TMM.2023.3321504}}

@article{KE2024110481,
title = {Text-based person search via cross-modal alignment learning},
journal = {Pattern Recognition},
volume = {152},
pages = {110481},
year = {2024},
issn = {0031-3203},
doi = {10.1016/j.patcog.2024.110481},
author = {Xiao Ke and Hao Liu and Peirong Xu and Xinru Lin and Wenzhong Guo},
}

@article{ZHANG2025111247,
title = {Local-enhanced representation for text-based person search},
journal = {Pattern Recognition},
volume = {161},
pages = {111247},
year = {2025},
issn = {0031-3203},
doi = {10.1016/j.patcog.2024.111247},
author = {Guoqing Zhang and Yuhao Chen and Yuhui Zheng and Gaven Martin and Ruili Wang},
}

@article{LIU2022108654,
title = {Making person search enjoy the merits of person re-identification},
journal = {Pattern Recognition},
volume = {127},
pages = {108654},
year = {2022},
issn = {0031-3203},
doi = {10.1016/j.patcog.2022.108654},
author = {Chuang Liu and Hua Yang and Qin Zhou and Shibao Zheng}
}

@article{LIU2023109636,
title = {BDNet: A BERT-based dual-path network for text-to-image cross-modal person re-identification},
journal = {Pattern Recognition},
volume = {141},
pages = {109636},
year = {2023},
issn = {0031-3203},
doi = {10.1016/j.patcog.2023.109636},
author = {Qiang Liu and Xiaohai He and Qizhi Teng and Linbo Qing and Honggang Chen}
}

@INPROCEEDINGS{RDE,
  author={Qin, Yang and Chen, Yingke and Peng, Dezhong and Peng, Xi and Zhou, Joey Tianyi and Hu, Peng},
  booktitle={CVPR}, 
  title={Noisy-Correspondence Learning for Text-to-Image Person Re-Identification}, 
  year={2024},
  volume={},
  number={},
  pages={27187-27196},
  doi={10.1109/CVPR52733.2024.02568}}

@article{KIM2026132885,
  title={{DiCo}: Disentangled Concept Representation for Text-to-Image Person Re-Identification},
  author={Kim, Giyeol and Eom, Chanho},
  journal={Neurocomputing},
  pages={132885. },
  year={2026},
  doi={10.1016/j.neucom.2026.132885},
  publisher={Elsevier}
}

@inproceedings{BAMG,
author = {Cheng, Keyang and Zou, Wenxuan and Gu, Hongjian and Ouyang, Anxiang},
title = {{BAMG}: Text-based Person Re-identification via Bottlenecks Attention and Masked Graph Modeling},
year = {2024},
isbn = {978-981-96-0965-9},
doi = {10.1007/978-981-96-0966-6_23},
pages={1809--1826},
booktitle = {ACCV}
}

@misc{nguyen2026itself,
  title        = {{ITSELF}: Attention Guided Fine-Grained Alignment for Vision--Language Retrieval},
  author       = {Nguyen, Tien-Huy and Tran, Huu-Loc and Ngo, Thanh Duc},
  year         = {2026},
  eprint       = {2601.01024},
  archivePrefix= {arXiv},
  primaryClass = {cs.CV}
}

@ARTICLE{samc,
  author={Lu, Zefeng and Lin, Ronghao and Hu, Haifeng},
  journal={IEEE Transactions on Information Forensics and Security}, 
  title={Mind the Inconsistent Semantics in Positive Pairs: Semantic Aligning and Multimodal Contrastive Learning for Text-Based Pedestrian Search}, 
  year={2024},
  volume={19},
  number={},
  pages={6409-6424},
  doi={10.1109/TIFS.2024.3417251}}

@article{AUL, title={Adaptive Uncertainty-Based Learning for Text-Based Person Retrieval}, volume={38}, DOI={10.1609/aaai.v38i4.28101}, abstractNote={Text-based person retrieval aims at retrieving a specific pedestrian image from a gallery based on textual descriptions. The primary challenge is how to overcome the inherent heterogeneous modality gap in the situation of significant intra-class variation and minimal inter-class variation. Existing approaches commonly employ vision-language pre-training or attention mechanisms to learn appropriate cross-modal alignments from noise inputs. Despite commendable progress, current methods inevitably suffer from two defects: 1) Matching ambiguity, which mainly derives from unreliable matching pairs; 2) One-sided cross-modal alignments, stemming from the absence of exploring one-to-many correspondence, i.e., coarse-grained semantic alignment. These critical issues significantly deteriorate retrieval performance. To this end, we propose a novel framework termed Adaptive Uncertainty-based Learning (AUL) for text-based person retrieval from the uncertainty perspective. Specifically, our AUL framework consists of three key components: 1) Uncertainty-aware Matching Filtration that leverages Subjective Logic to effectively mitigate the disturbance of unreliable matching pairs and select high-confidence cross-modal matches for training; 2) Uncertainty-based Alignment Refinement, which not only simulates coarse-grained alignments by constructing uncertainty representations but also performs progressive learning to incorporate coarse- and fine-grained alignments properly; 3) Cross-modal Masked Modeling that aims at exploring more comprehensive relations between vision and language. Extensive experiments demonstrate that our AUL method consistently achieves state-of-the-art performance on three benchmark datasets in supervised, weakly supervised, and domain generalization settings. Our code is available at https://github.com/CFM-MSG/Code-AUL.}, number={4}, journal={AAAI}, author={Li, Shenshen and He, Chen and Xu, Xing and Shen, Fumin and Yang, Yang and Shen, Heng Tao}, year={2024}, month={Mar.}, pages={3172-3180} }
\end{document}